\documentclass{article}

\usepackage{arxiv}

\usepackage[utf8]{inputenc} 
\usepackage[T1]{fontenc}    
\usepackage{hyperref}       
\usepackage{url}            
\usepackage{booktabs}       
\usepackage{amsfonts}       
\usepackage{nicefrac}       
\usepackage{microtype}      
\usepackage{lipsum}		
\usepackage{graphicx}
\usepackage[numbers]{natbib}
\usepackage{doi}

\usepackage{url}
\usepackage{tabulary}
\usepackage{svg}
\usepackage{soul}
\usepackage{amsmath,amssymb}
\usepackage[none]{hyphenat}
\usepackage{xurl} 
\usepackage{caption}
\usepackage{subcaption}
\usepackage{comment}
\usepackage{tablefootnote}
\usepackage{svg}
\usepackage{academicons}
\usepackage{bm}

\usepackage{bbding}

\graphicspath{{images/}}

\DeclareUnicodeCharacter{0141}{\L{}}
\DeclareUnicodeCharacter{0142}{\l{}}

\newcommand{\review}[1]{{#1}}

\usepackage{bbding}
\newenvironment{minipeqn}[1][]{\begin{minipage}[t][#1]{.45\textwidth}\begin{equation}}{\end{equation}\end{minipage}}

\title{Secure Federated Learning for Residential Short Term Load Forecasting}

\author{ \href{https://orcid.org/0000-0003-1326-6134}{\includegraphics[scale=0.06]{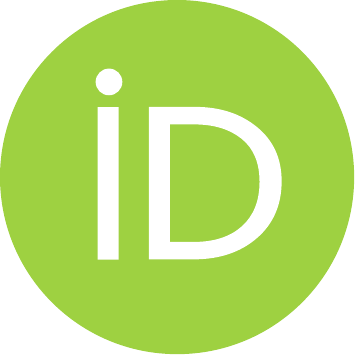}\hspace{1mm}Joaquin Delgado Fernandez} \\
	Interdisciplinary Centre for\\ Security, Reliability and Trust (SnT)\\
	FINATRAX, University of Luxembourg\\
	Luxembourg \\
	\texttt{joaquin.delgadofernandez@uni.lu} \\
	\And
	\href{https://orcid.org/0000-0002-9032-7183}{\includegraphics[scale=0.06]{orcid.pdf}\hspace{1mm}Sergio Potenciano Menci} \\
	Interdisciplinary Centre for\\ Security, Reliability and Trust (SnT)\\
	FINATRAX, University of Luxembourg\\
	Luxembourg \\
	\texttt{sergio.potenciano-menci@uni.lu} \\
	\AND
	\href{https://orcid.org/0000-0002-8677-3463}{\includegraphics[scale=0.06]{orcid.pdf}\hspace{1mm}Charles Lee} \\
	Interdisciplinary Centre for\\ Security, Reliability and Trust (SnT)\\
	FINATRAX, University of Luxembourg\\
	Luxembourg \\
	\texttt{charles.lee@uni.lu} \\
	\And
	\href{https://orcid.org/0000-0001-7996-4678}{\includegraphics[scale=0.06]{orcid.pdf}\hspace{1mm}Alexander Rieger} \\
	Interdisciplinary Centre for\\ Security, Reliability and Trust (SnT)\\
	FINATRAX, University of Luxembourg\\
	Luxembourg \\
	\texttt{alexander.rieger@uni.lu} \\
	\AND
	\href{https://orcid.org/00000-0001-7037-4807}{\includegraphics[scale=0.06]{orcid.pdf}\hspace{1mm}Gilbert Fridgen} \\
	Interdisciplinary Centre for\\ Security, Reliability and Trust (SnT)\\
	FINATRAX, University of Luxembourg\\
	Luxembourg \\
	\texttt{gilbert.fridgen@uni.lu} \\
}

\renewcommand{\shorttitle}{Privacy-preserving Learning for residential STLF}

\hypersetup{
pdftitle={Privacy-preserving Federated Learning for Residential Short Term Load Forecasting},
pdfsubject={cs.AI},
pdfauthor={Joaquin Delgado Fernandez, Sergio Potenciano Menci, Charles Lee, Gilbert Fridgen},
pdfkeywords={Deep neural networks,Differential privacy, Federated learning, Secure aggregation, Privacy-preserving federated learning, Short-term load forecasting},
}

\begin{document}
\maketitle

\begin{abstract}
With high levels of intermittent power generation and dynamic demand patterns, accurate forecasts for residential loads have become essential. Smart meters can play an important role when making these forecasts as they provide detailed load data. However, using smart meter data for load forecasting is challenging due to data privacy requirements. This paper investigates how these requirements can be addressed through a combination of federated learning and privacy preserving techniques such as differential privacy and secure aggregation. For our analysis, we employ a large set of residential load data and simulate how different federated learning models and privacy preserving techniques affect performance and privacy. Our simulations reveal that combining federated learning and privacy preserving techniques can secure both high forecasting accuracy and near-complete privacy. Specifically, we find that such combinations enable a high level of information sharing while ensuring privacy of both the processed load data and forecasting models. Moreover, we identify and discuss challenges of applying federated learning, differential privacy and secure aggregation for residential short-term load forecasting.

\end{abstract}


\sloppy

\section{Introduction}
\label{sec:introduction}


As the supply from intermittent and difficult-to-forecast renewable power sources increases, load forecasting – and especially residential short-term load forecasting (STLF) - is becoming ever more crucial for the reliability of modern power systems \cite{forecasting_importance, PETROPOULOS2022}. Residential STLF covers forecasting windows from a few minutes to a week ahead \cite{PETROPOULOS2022, etnso_forecasting_def}. It plays an important role for many operational processes in the power system, such as planning, operating, and scheduling \cite{Antonio_comillas, LUSIS2017654}. For instance, it enables energy providers to identify gaps between supply and demand in their customer portfolios. These gaps typically lead to high imbalance costs and ultimately to higher electricity prices for residential customers \cite{energy_supply_costs, energy_supply_2.0}. 

Traditionally, residential STLF has relied on aggregated load data and reference load profiles \cite{LUSIS2017654, MALTAIS2021118229, smart_meter_analytics}. Yet, aggregation and reference profiles are often ill-suited for power systems with a high share of distributed generation and active demand-side management \cite{LUSIS2017654, MALTAIS2021118229}. Moreover, they have become less reliable with residential heating and mobility being increasingly electric \cite{electrification, iea_consumption} and consumption patterns growing more dynamic, for instance, due to fluctuating levels of remote work \cite{en14040980}. These trends make accurate forecasting of individual residential loads an important priority.

There are various traditional methods for more granular STLF, but most build on limiting linearity assumptions (correlation between values and past values) even though residential load patterns are often highly dynamic \cite{LUSIS2017654}. Examples include time series models that rely on seasonal autoregressive integrated moving averages (ARIMA) \cite{LUSIS2017654, kaur2017timeARIMA}, exponential smoothing, or linear transfer functions. Residential STFL is thus increasingly relying on methods that can work with non-linear dependencies, such as many Artificial Intelligence (AI) models ~\cite{s20051399, forecasting_approaches, Electric_load_forecasting, 910780,ardabili2019advances}. 

A core challenge for any of these methods is the availability of granular data \cite{problem_data_forecasting}. In many countries, this 'data scarcity' problem is tackled by pushing for advanced metering infrastructure (AMI), which substantially increases the resolution of residential load data \cite{smart_meters_survey}. STLF methods can make use of this data using either 'centralized' or 'decentralized' approaches. Centralized approaches transfer smart meter data to a central forecasting system. While these forecasting systems promise very accurate results, they face a twofold problem. First, they are subject to substantial privacy challenges because smart meter data are often easily attributable to natural persons. That is, data collected from smart meters can be detailed enough to permit the identification of specific customers \cite{hinterstocker2017disaggregation}. The transfer and aggregation of smart meter data is thus typically subject to data privacy regulations such as the European Union's General Data Protection Regulation (GDPR) and its obligations and requirements for processing personal data~\cite{MCKENNA2012807, smart_meter_eu_project_data_access}. Second, there are considerable regulatory uncertainties. In particular, it is often unclear how device ownership (who owns the smart meter) and aggregation impact data ownership. Moreover, specific regulations for smart meter data are typically absent~\cite{smart_meter_ownership, smart_meters}. These regulatory uncertainties often mean that centralized approaches such as Belgium's Atrias \cite{atrias_2021}, or Norway's Elhub \cite{elhub_2021}, which provide so-called \textit{data lakes}, may not be desirable.

Decentralized approaches aim to tackle some of these issues by processing smart meter data locally. A particularly promising of these decentralized approaches is Federated Learning (FL)~\cite{mcmahan2017communicationefficient, konecny2016federated}. Federated Learning is a machine learning technique that offers a collaboration framework for clients. In a so-called 'federation' clients jointly train and share prediction models instead of training data. Although FL cannot guarantee privacy by itself \cite{zhu2019deep, geiping2020inverting}, it can be combined with privacy-preserving techniques such as differential privacy (DP) and secure aggregation (SecAgg).

Even though such a combination could substantially benefit residential STLF, academic attention to FL has been limited so far \cite{briggs2021federated, 9469923, He2021ShortTermRL, taik2020electrical, Biswal2021AMIFMLAP, Li2020FederatedLU, Xu2021LSTMSR, FEKRI2021107669, Shi2022DeepFA, Lin2022PrivacyPreservingHC, Husnoo2022FedREPTH, Khalil2021FederatedLF} and the two components have mostly been considered mostly in isolation \cite{BARBOSA2016355, dp_smart_meter, dp_imperial_colleague}. With this paper, we seek to close several gaps in the literature on FL-based STLF: Firstly, we aim to deepen the understanding of FL-based STLF by examining the effects of clustering based on Pearson correlation and the effects of architectural complexity. Secondly, we analyze the privacy and performance effects of adding privacy-preserving techniques (DP and SecAgg) to FL. Third, we identify key challenges associated with using a combination of FL and privacy-preserving techniques.

To do so, we conduct the following analysis: Initially, we identify promising NN architectures from a review of the recent FL literature. Subsequently, we select the most effective of these architectures and investigate six scenarios using real-world historical data. In a first scenario, we evaluate the performance of the selected architecture in a 'centralized' setting to establish a performance benchmark for the remaining five FL scenarios. In the second scenario, we investigate the performance and computational cost effects of moving from a centralized setting to a FL setting. In a third scenario, we then examine the effects of using correlated training data based on Pearson correlation and socio-economic factors. Correlation is typically avoided in non-federated ML models to increase data variability. Yet, for FL models, correlated data may increase forecasting accuracy \cite{taik2020electrical} and mitigate problems with non-IID (non-independent and non-identically distributed) data. In the fourth scenario, we reflect on the trend to work with ever more complex models and explore the effects of increasing the complexity of the NN's architecture. In scenarios 5 and 6, we study how privacy-preserving techniques affect the training and performance of federated models. Specifically, we investigate the effect of different DP implementations (i.e., clipping techniques) and SecAgg on accuracy, privacy, and computational costs.

The remainder of the paper is structured as follows. Section~\ref{sec:litrature} provides an overview of related work on the use of NNs for STLF, FL, and privacy-preserving techniques. Section~\ref{sec:simulation} covers our evaluation method, including the simulation environment, dataset and evaluation metrics. Section \ref{sec:evaluation_design} describes our evaluation design. It covers the selection of the baseline NN architecture, the specification of the analyzed differential privacy and secure aggregation techniques, the training process for the federated learning models, and the design of six evaluation scenarios. Section~\ref{sec:evaluation} presents the evaluation results for the six scenarios. Finally, section~\ref{sec:conclusions} provides a synthesis of our results and points out directions for further research.

\section{Related work}
\label{sec:litrature}

\subsection{Federated learning}
\label{sec:FL}

In most fields, AI-based methods have already proven their value. However, their performance is highly dependent on the quantity and quality of available training data. Generally speaking, AI-based methods are typically limited by data fragmentation and isolation – mostly due to competitive pressure and tight regulatory frameworks (related to data privacy and security). To address these challenges, McMahan et al. proposed a new technique, FL \cite{mcmahan2017communicationefficient, konecny2016federated}. The main idea of FL is to collaboratively train machine learning models between multiple independent clients without moving or revealing the training data. In other words, FL allows competing participants to leverage each others' datasets without revealing their own individual datasets. In doing so, models trained with FL enable more accurate forecasts than models that were independently trained by each client. To date, there are two canonical training algorithms for FL and four different configurations for the distribution of data and errors.

The two canonical training algorithms are: federated stochastic gradient descent (Fed-SGD) and federated averaging (Fed-Avg) \cite{mcmahan2017communicationefficient}. Fed-SGD works by averaging the client's gradients after every pass through a local data batch. More specifically, Fed-SGD clients compute gradients of their 'loss' for a sub-set of their data. The loss is a non-parametric function that penalizes bad predictions and to minimize it, the clients need to move toward the empirical minimum by taking steps in the opposite direction of the gradient. Clients subsequently send their locally computed gradients to a central server. The central server aggregates and averages them - either equally or in a weighted manner - to update the model weights. These updated weights are again sent to the clients and each client trains their local model with the updated weights. Training continues in an iterative manner until a pre-defined number of so called communication rounds have been reached or a common goal is achieved. In Fed-SGD, a communication round represents a full pass through all batches.

In Fed-Avg, the clients send their model weights instead of their gradients.
 Once the central server has received the weights, it aggregates and averages them to arrive at a new 'consensus' that will be sent back to the clients for the next training round. Unlike Fed-SGD, Fed-Avg does not split the training data into batches, which has two effects: the number of communication rounds is reduced substantially (only once per epoch) and an improvement in forecasting accuracy ~\cite{mcmahan2017communicationefficient,FEKRI2021107669}. As in Fed-SGD, the training process continues until the pre-defined number of epochs has been reached or a common goal is achieved.

Besides different algorithms, FL applications can also differ in their configurations. These configurations depend on how the data is structured. More specifically, they depend on the configuration of the feature space $\mathcal{X}$, the label space $\mathcal{Y}$, and the space formed by the identifiers $\mathcal{I}$. Different setups of the triplet ($\mathcal{X}, \mathcal{Y},\mathcal{I}$) can be classified as Horizontal, Vertical, Transfer and Assisted Federated Learning ~\cite{yang2019federated}. 
Take for instance two clients $i$ and $j$.
\begin{itemize}
    \item Horizontal Federated Learning is when $i$ and $j$ share the same feature space such that $\mathcal{X}_i = \mathcal{X}_j$ but their label spaces $\mathcal{Y}$ are different so that $\mathcal{Y}_i \neq \mathcal{Y}_j$. In our residential STLF example, Horizontal FL would be applicable when the model is to be trained on smart meter data from a range of clients with the same feature set (consumption, weather profile, etc.) and the data is held by different companies.
    \item Vertical Federated Learning is when $\mathcal{I}_i = \mathcal{I}_j$, but $\mathcal{X}_i \neq \mathcal{X}_j$ and $\mathcal{Y}_i \neq \mathcal{Y}_j$. This would be the case, for instance, when two companies have access to the same client but each of them holds a different feature set regarding the client.
    \item Federated Transfer Learning happens when $\mathcal{X}_i \neq \mathcal{X}_j$, $\mathcal{Y}_i \neq \mathcal{Y}_j$, $\mathcal{I}_i \neq \mathcal{I}_j,\;  \forall \mathcal{D}_i,\mathcal{D}_j, i \neq j$. Federated Transfer Learning can be used, for instance, when two companies have different clients and feature sets but want to nevertheless collaboratively train a model.
    \item  Assisted Learning (AL) is done through collided data between clients. Xian et al. \cite{xian2020assisted} define collision as when clients with the same data entries of a dataset $\mathcal{D}$ have different feature spaces $ \mathcal{I}_i = \mathcal{I}_j , \mathcal{X}_i \neq \mathcal{X}_i \; \forall \mathcal{D}_i,\mathcal{D}_j, i \neq j$. One client may use the errors of another for their own benefit by increasing their training performance. 
\end{itemize}

Regardless of the chosen algorithm and configuration, FL is vulnerable to moral hazard~\cite{DBLP:journals/corr/abs-1912-04977} or so-called 'soft' attacks on the contextual integrity of the shared data. Moral hazard arises because FL is by nature collaborative~\cite{DBLP:journals/corr/McMahanMRA16}. Multiple clients must work together to train models iteratively using the respective data at their disposal. If one or several of these clients manipulate the joint training process, it does not work. In effect, federated learning requires trust between the clients involved.

\subsection{FL-based short term Load forecasting}
\label{sec:STLF}

Short-term load forecasting is a complex, multivariate time series problem. Its complexity is high because residential load data is often replete with irregularities, missing or inaccurate values, and seasonality. Petropoulus et al. \cite{PETROPOULOS2022} provide an in-depth overview of these challenges. Yet, they also point out the increasing importance and momentum that STLF has gained over recent years. STLF is crucial because system operators require it for unit commitment and optimal power flow calculations \cite{PETROPOULOS2022, Antonio_comillas, LI2020114850}. Moreover, it enables utilities, energy suppliers, and distribution grid operators (DSOs) to optimize their customer portfolios, design tariffs, and strategically adapt flexibility offerings \cite{PETROPOULOS2022, Antonio_comillas}.

STLF typically build on three groups of methods: traditional methods, AI-based methods, and hybrid methods that integrate traditional and AI-based components \cite{PETROPOULOS2022}. Traditional methods such as ARIMA can capture seasonal trends but fall short when it comes to non-linear patterns and non-aggregated data. At the same time, they are simple to use and have light computational costs \cite{PETROPOULOS2022}. AI-based methods, in turn, are well suited to identifying non-linear patterns and work well with individual (i.e., residential level) and aggregated data (i.e., substation level) \cite{LUSIS2017654,Vos2018}.

Within the larger group of AI-based methods, FL is a relatively new but increasingly popular method for STLF. Our following overview of these FL studies which follows is based on a search in Semantic Scholar using the following search terms: \textit{short-term load forecasting neural networks} and \textit{Federated Learning for Residential Short Term Load Forecasting}.

The first group of studies employ Fed-SGD \cite{He2021ShortTermRL, Lin2022PrivacyPreservingHC}. He et al. \cite{He2021ShortTermRL} additionally use k-means clustering and compare performance between six scenarios with a different number of clusters in each scenario. Their results suggest that grouping data based on comparable load patterns substantially improves the performance of FL models. Lin et al. \cite{Lin2022PrivacyPreservingHC}, in turn, focus on limiting the high computational cost of Fed-SGD. To this end, they introduce an asynchronous stochastic gradient descent algorithm with delay computation (ASGD-DC). Specifically, their algorithm uses a Taylor expansion to compensate for the delay of clients with lower computational power. 

The second and substantially larger group of studies employ Fed-Avg.
Similar to He et al. \cite{He2021ShortTermRL}, Briggs et al. \cite{briggs2021federated}, Savi et al. \cite{9469923}, Afaf et al. \cite{taik2020electrical}, and Biswal et al. \cite{Biswal2021AMIFMLAP} investigate different forms of clustering for Fed-Avg. Their findings suggests that clustering based on k-means and socio-economic factors can also substantially improve the performance of Fed-Avg. With certain caveats, their findings also suggest that its possible to train good models with a small number of clients. Li et al. \cite{Li2020FederatedLU}, in turn, use Fed-Avg to compare the effects of different federation sizes, ranging the number of clients from 2, to 4, and 6. They also vary the number of training rounds (epochs) from 5 to 15. Their results suggest performance is increased by increasing the number of clients and training rounds.

Xu et al. \cite{Xu2021LSTMSR} as well as Husnoo et al. \cite{Husnoo2022FedREPTH} investigate the effect of increasing the number of clients participating in the training rounds. Their results show a considerably drop in performance for the higher participation cases. This drop appears to be the result of non-IDD consumption data between the clients. 

Khalil et al. in \cite{Khalil2021FederatedLF} use Fed-Avg to train a FL model for building control, replicating the use of FL for household training. They consider six floors of a seven-story building as clients. They later personalize the global FL model for the 7th floor - not used in the FL training - by running locally five additional rounds (epochs) and not sharing the data with the global model. Their results suggest that even the personalized FL model can help a smart building controller reduce total electricity consumption using FL.

In terms of relative performance, Fekri et al. \cite{FEKRI2021107669} find that Fed-Avg provides more accurate results for STLF than Fed-SGD. Shi et al. \cite{Shi2022DeepFA},in turn, look beyond canonical FL and use a multiple kernel variant of maximum mean discrepancies (MK-MMD) to fine-tune the central server model (global). They train for several rounds using transfer learning to adapt the global model to specific customers. Their results indicate better performance than a canonical Fed-avg implementation. 

The works of \cite{briggs2021federated, 9469923, He2021ShortTermRL, taik2020electrical, Biswal2021AMIFMLAP, Li2020FederatedLU, Xu2021LSTMSR, FEKRI2021107669, Shi2022DeepFA, Lin2022PrivacyPreservingHC, Husnoo2022FedREPTH, Khalil2021FederatedLF} provide important stepping-stones in FL-based STLF. In particular, they clearly indicate the prospect of using collaborative training to create accurate forecasting models. However, they provide only limited insights into the challenges of using FL. In particular, it is not yet clear if different but simpler clustering techniques such as Pearson correlation are also effective. Also, prior literature has not yet looked at the effect of architectural complexity. Moreover, existing studies do not or only in a very limited way account for matters of privacy. Thus, this paper aims to provide a better understanding of clustering and architectural complexity and explores the addition of different privacy preserving techniques.

\subsection{NN architectures for FL-based short term Load forecasting}
\label{sec:NN}

The studies presented on FL-based STLF use a range of different NN architectures (Table \ref{tab:forecasting_literature}). Overall, the architectures have become deeper (i.e., multi-layered) over time as depth is typically associated with more accurate results \cite{Vos2018}. In terms of layer design, we found Fully Connected layers (FCL), Long Short-term Memory (LSTM) Layers \cite{hochreiter1997long} and Convolutional Neural Networks (CNN). LSTMs have feedback connections which understand the dependence between items in a sequence and which make them suitable for temporal pattern recognition. CNN layers emulate human retinas and can capture the spatial distribution of graphic patterns. Moreover, we found Encoder-Decoder or autoencoder architectures \cite{marino2016building}. In these architectures, the NN is provided with a sequence (a vector) as an input and maps this sequence to another sequence. Encoder-Decoder architectures reduce the effects of outliers because they transpose the original input space into a differently encoded space~\cite{sutskever2014sequence, cho2014properties}. Sehovac et al. \cite{Sehovac_NN_complex} present a particular interesting example of a Seq2Seq architecture that includes an attention mechanism to help the decoder extract additional information.

Aside from different layer designs, we also identified hybrid designs. For instance, Kim et al. \cite{kim2019predicting} use CNN with LSTM layers to find both spatial and temporal patterns. Building on their work, Tuong et al. \cite{app9204237} add a bi-directional LSTM layer to identify temporal trends both forward and backwards in time. Similarly, Zulfiqar Ahmad et al. \cite{s20051399} combine Seq2Seq from \cite{marino2016building} with a CNN layer design. This combination allows for the capture of both temporal and spatial patterns and offers protection against outliers. Shi et al. \cite{7885096} take a different path by clustering and pooling the training data to increase variability and reduce overfitting. 

\newpage

\begin{table}[h!]

\caption{Neural network architectures for FL-based and non FL-based STLF. }
\label{tab:forecasting_literature}

\resizebox{0.95\textwidth}{!}{%
\begin{tabular}{m{0.2\textwidth}m{0.4\textwidth}m{0.35\textwidth}m{0.05\textwidth}}
\toprule
    \textbf{Method} & \textbf{Dataset} & \textbf{Neural Network Architecture} & \textbf{Year}\\ \midrule
    
    Marino et al. \hspace{7.5pt} \cite{marino2016building} &
    UCI - Individual household electric power consumption &
    LSTM + Repeat vector + LSTM + 2x FCL &
    2016 \\
    Kong et al.   \hspace{15.5pt} \cite{8039509} &
    Australia SGDS Smart Grid Dataset &
    Stacked LSTM + FCL &
    2017 \\
    Li et al.     \hspace{28.5pt} \cite{en10101525} &
    Fremont, CA 15min Retail building electricity load &
    Missing or incomplete architecture description &
    2017 \\
    Shi et al.    \hspace{27pt}\cite{7885096} &
    Irish CBTs - Residential and SMEs &
    Stacked LSTM + Pooling mechanism &
    2018 \\
    Yan et al.   \hspace{22pt} \cite{en11113089} &
    UK-DALE Domestic Appliance-Level Electricity dataset &
    2x Conv + 1x LSTM + FCL &
    2018 \\
    Kim and Cho   \hspace{7pt} \cite{en12040739} &
    UCI - Individual household electric power consumption &
    Missing or incomplete architecture description &
    2019 \\
    Kim and Cho   \hspace{7pt} \cite{kim2019predicting} &
    UCI - Individual household electric power consumption &
    2x Conv + LSTM + 2x FCL &
    2019 \\
    Le et al.    \hspace{27pt} \cite{app9204237} &
    UCI - Individual household electric power consumption &
    2x Conv + Bi + LSTM + 2x FCL &
    2019 \\
    Khan et al. \hspace{16pt} \cite{s20051399} &
    UCI - Individual household electric power consumption &
    2x Conv + 2x LSTM (Encoder) + 2x LSTM (Decoder) + 2x FCL &
    2020 \\
    Afaf et al. \cite{taik2020electrical} & 
    Pecan Street Research Institute &
    2x LSTM (same size) + FCL &
    2020 \\
    Sehovac et al. \cite{Sehovac_NN_complex} & 
    Non-disclosed or private data &
    Sequence to Sequence with attention &
    2020 \\
    \review{ Li et al. \cite{Li2020FederatedLU}} &
    \review{ Global Energy Forecasting Competition 2012} &
    \review{ Missing or incomplete architecture description} &
    \review{ 2020} \\
    \review{ Xu et al. \cite{Xu2021LSTMSR}} &
    \review{ Pecan Street Research Institute} &
    \review{ Missing or incomplete architecture description} &
    \review{ 2021} \\
    \review{ Briggs et al. \cite{Briggs2021FederatedLF_idea_not_full_paper}} &
    \review{ Low Carbon London Dataset} &
    \review{ 2x LSTM (same size) + FCL} &
    \review{ 2021} \\
    \review{ He et al. \cite{He2021ShortTermRL}} &
    \review{ Australia SGDS Smart Grid Dataset} &
    \review{ 2x LSTM (same size) + FCL} &
    \review{ 2021} \\
    \review{ Savi et al. \cite{9469923}} &
    \review{ Low Carbon London Dataset} &
    \review{ LSTM (64) + LSTM (32) + FCL} &
    \review{ 2021} \\
    \review{ Zhao et al. \cite{Zhao2021ADP}} &
    \review{ Pecan Street Research Institute} &
    \review{ 2x LSTM (same size) + FCL} &
    \review{ 2021} \\
    \review{ Biswal et al. \cite{Biswal2021AMIFMLAP}} &
    \review{ Commission for Energy Regulation (CER)} &
    \review{ Missing or incomplete architecture description} &
    \review{ 2021} \\
    \review{ Khalil et al. \cite{Khalil2021FederatedLF}} &
    \review{ CU-BEMS, smart building electricity consumption and indoor environmental sensor datasets} &
    \review{ Missing or incomplete architecture description} &
    \review{ 2021} \\
    \review{ Shi et al. \cite{Shi2022DeepFA}} &
    \review{ Low Carbon London Dataset} &
    \review{ Missing or incomplete architecture description} &
    \review{ 2022} \\
    \review{ Lin et al. \cite{Lin2022PrivacyPreservingHC}} &
    \review{ Commission for Energy Regulation (CER)} &
    \review{ Missing or incomplete architecture description} &
    \review{ 2022} \\
    \review{ Husnoo et al. \cite{Husnoo2022FedREPTH}} &
    \review{ Solar Home Electricity Data from Eastern Australia} &
    \review{ LSTM (256) + LSTM (128) + FCL} &
    \review{ 2022} \\

\bottomrule
\end{tabular}%
}
\end{table}

\subsection{Privacy preserving techniques for federated learning}
\label{sec:PPT}

Privacy-preserving techniques can support the design of forecasting systems that comply with privacy requirements and regulations \cite{MCKENNA2012807, Bennett_2018_privacy_importance, FL_survey}. From an organizational perspective, these techniques allow competing agents like energy providers to cooperate and integrate with utilities and DSOs \cite{smart_meter_eu_project_data_access, Bennett_2018_privacy_importance}. Furthermore, their use might facilitate the creation of local markets that support the energy transition \cite{PRESSMAIR2021128323}. 

Privacy-preserving techniques are especially relevant for FL. Although FL offers considerable improvements over centralized ML methods, it does not guarantee privacy. Firstly, the shared data (gradients or model weights) may allow inadvertent attribution, and secondly, privacy can be compromised through the communication between clients and the central server. For instance, Zhu et al. found a way to use gradient updates to reconstruct the training data of a client \cite{zhu2019deep}. This effectively means that gradient updates are to be treated as personal data and that FL requires additional measures when data privacy is required. In the following, we describe two such measures: DP as a way to anonymize training data and SecAgg as a mechanism to enable privacy-sensitive communication between clients and the central server. 

Dwork \cite{Dwork06} introduces DP as a technique to guarantee privacy when retrieving information from a dataset. As described in \cite{dwork2014algorithmic}, "differential privacy addresses the paradox of knowing nothing about an individual while learning useful information about a population." DP hides individual data trends by using additive noise. In more technical terms, Dwork \cite{Dwork06} introduced epsilon differential privacy ($\epsilon$-DP) as follows: \textit{"For every pair of inputs $x$ and $y$ that differ in one row, for every output in S, an adversary should not be able to use the output in S to distinguish between any $x$ and $y$"}. The privacy budget ($\epsilon$) determines how much of an individual's privacy a query may use, or to what extent it may increase the risk of breaching an individual's privacy. A value of $\epsilon = 0$ represents perfect privacy, which means that privacy cannot be compromised through any analysis on a dataset in question \cite{DP_non_tech}. Jayaraman et al. \cite{jayaraman2019evaluating} extended the concept of ($\epsilon$-DP) to ($\epsilon,\delta$-DP) where $\delta$ is the failure probability to better control for the tails of the privacy budget.

DP is typically implemented by adding random noise to data queries. This noise is usually sampled from a Laplacian or Gaussian distribution~\cite{dwork2014algorithmic}. Finding an adequate noise level is crucial but not trivial - especially for FL. Too much noise can not only hide patterns in the data but also complicate convergence of the local models due to the random updates of the patterns during training. Simply speaking, more noise means more privacy, but more noise also means less accuracy. 

An alternative to adding noise to the training process or the data is using secure multi-party computation (SMPC) protocols, which enable privacy-preserving communication. One such protocol is SecAgg \cite{SecAgg}. SecAgg uses cryptographic primitives that prevent the central server from reconstructing each client's involvement and contribution. In more technical terms, SecAgg allows a set of distributed, unknown clients to aggregate a value $x$ without revealing the value to the other clients. The backbone of SecAgg is Shamir's \textit{t-out-of-n} Secret Sharing. It enables a user to split a secret $s$ into $n$ shares~\cite{shamir1979share}. To reconstruct the secret, more than $t-1$ shares are needed to retrieve the original secret $s$. Any allocation with less than $t-1$ shares will provide no information about the original secret. SecAgg implies two main algorithms: sharing and reconstruction. The sharing algorithm transforms a secret into a set of shares of the secret that are each associated with a client. Following \cite{shamir1979share}, these shares are constructed in such a way that collusion between $t-1$ participants ($t$ being the total number of participants) is insufficient to disclose other clients' private information. The reconstruction algorithm works in the opposite direction. It takes the mentioned shares from the clients and reconstructs the shared secret.

Of the two privacy-preserving techniques, only DP has so far been examined in the context of residential STLF. Chhachhi et all. \cite{dp_imperial_colleague}, Eibl et al. \cite{eibl2017differential}, and Zhao et al. \cite{6847974} use DP to train a 'centralized' machine learning model. More specifically, they perturb the datasets by adding noise drawn from either a Gaussian or Laplacian distribution before each training round of the model. To the best of our knowledge, Zhao et al. \cite{Zhao2021ADP} are the first to combine FL and DP for STLF. Specifically, they include DP in the training process of a Fed-Avg  model. However, they do not systematically analyze different DP parameters. Moreover, they do not look at secure multi-party computation protocols, such as SecAgg.



\section{Method}
\label{sec:simulation}

\subsection{Simulation environment}
\label{subsec:simulation_environment}
The evaluations in this paper are based on simulations we ran on the IRIS Cluster of the high performance computer (HPC) facilities of the University of Luxembourg~\cite{HPC}. The simulations ran in an environment with 32 Intel Skylake cores and two NVIDIA Tesla V100 with 16GB or 32GV depending on the allocation. We programmed the federation code in Python and based it on the machine learning framework provided by Tensorflow-Federated~\footnote{https://github.com/tensorflow/federated} (TFF). The DL models are written in Keras \cite{chollet2015keras}. 

\subsection{Dataset}
\label{subsec:dataset}
For our simulations, we used a large dataset from the Low Carbon London project, which was conducted by UK Power Networks between November 2011 and February 2014 in London, United Kingdom (herein LCL dataset) \cite{uk_data}. It contains the electrical consumption\,[kWh] data from 5567 households in a half-an hour resolution. The LCL dataset also contains a socio-technical classification of the households following the ACORN scheme \cite{acorn} and is divided into individual household entries known as LCLid (Low Carbon London id). 

\begin{figure}[ht!]
\centering
\includegraphics[width=.8\columnwidth]{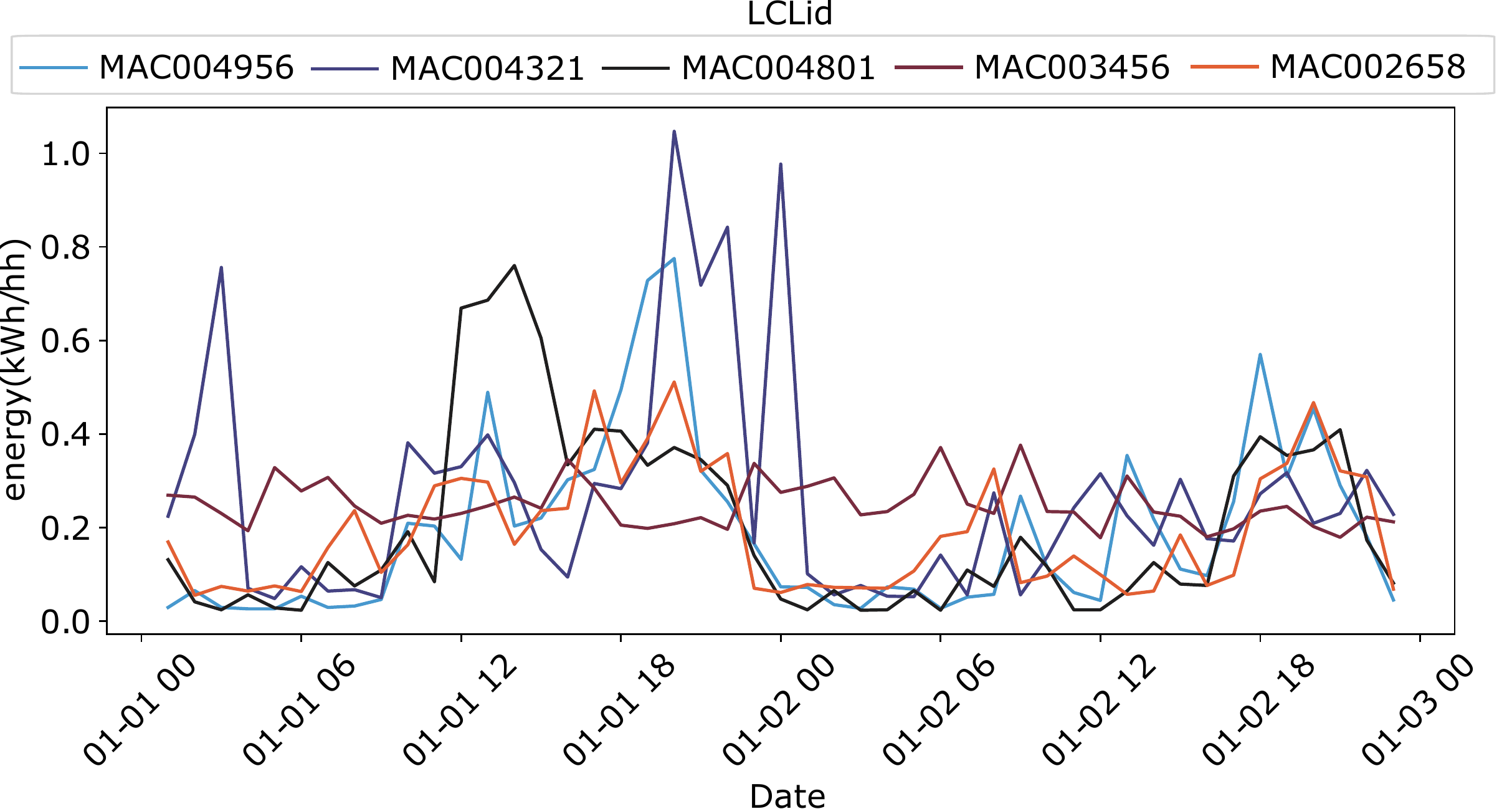}
\caption{Energy consumption (kWh/h) of 4 LCLIds  from 01 January 2013 to 03 January 2013. \label{4lclids}}
\end{figure}

To make the dataset ready for our simulations, we treated it in a \review{4}-step procedure. First, we reduced the resolution of the LCL dataset to hourly values. The down-scaled values in the treated data set are the sum of two subsequent half-hour values in the original data set. This treatment significantly reduced the computational burden of our simulations. Secondly, we trimmed outliers or null values. Thirdly, we scaled all variables to have the same range using a Min-Max scaler. This re-scaling was necessary to ease the FL learning process as all values have to be in a known range, in our case: 0 to 1. Fourthly and finally, we split the dataset into a training and validation dataset. The training dataset (75\%) contains electrical consumption data from January to December 2013 and the validation set (25\%) covers data from January 2014 to March 2014. In Figure~\ref{4lclids}, we provide an example of the processed data. It visualizes the electricity consumption\,[kWh] of 5 randomly selected households for a 2 day period using 1h timestamps.

\subsection{Evaluation metrics}
\label{subsec:metrics}

Evaluation metrics offer an important means for the training and testing of forecasting models. However, the use of certain metrics can lead to undesirable results because FL models are known to converge to a \textit{middle point} \cite{Li_2020}. More specifically, FL models optimize the error of prediction with respect to the ground truth. In a distributed environment where there are \textit{many such truths}, the models tend to minimize the mean of the loss across datasets. This tendency can provoke FL models to predict the average of each of the datasets and hence offer promising mean squared errors (MSE, Equation~\ref{eq:MSE}) and mean absolute errors (MAE, Equation~\ref{eq:MAE}). Such predictions, however, mean that the FL model did not learn local patterns in the data.

Therefore, MSE and MAE are typically not enough to evaluate the performance of a FL model and additional metrics, such as mean absolute percentage error (MAPE, Equation~\ref{eq:MAPE}) and root mean square error (RMSE, Equation~\ref{eq:RMSE}), are needed to quantify deviations of model predictions from the ground truths. The formal equations for these four metrics are as follows:

\noindent\begin{tabular*}{\textwidth}{ll}
 \begin{minipeqn}
    \label{eq:MSE}
    MSE = \frac{1}{n}\sum_{i=1}^{n}\left ( y_{i} - x_{i} \right )^2
\end{minipeqn}&
\begin{minipeqn}[]
    \label{eq:MAE}
    MAE = \frac{1}{n}\sum_{i=1}^{n}\left | y_{i} - x_{i} \right |
\end{minipeqn}\\
\begin{minipeqn}
 \label{eq:MAPE}
    MAPE = \frac{100}{n}\sum_{t=1}^n \left | \frac{x_{i}-y_{i}}{x_{i}} \right |
\end{minipeqn}&
 \begin{minipeqn}[]
   \label{eq:RMSE}
    RMSE = \sqrt{\left(\frac{1}{n}\right)\sum_{i=1}^{n}(y_{i} - x_{i})^{2}}
\end{minipeqn}\\
\end{tabular*}

\section{Evaluation}
\label{sec:evaluation_design}

\subsection{Selection of a baseline neural network architecture}
\label{subsec:DLArchitecture}

One crucial aspect for any AI method and specifically FL is the selection of the underlying NN architecture. To pick an architecture for our evaluation, we compared those in Table \ref{tab:forecasting_literature} that had a clear ‘implementation guide’ we could replicate. For this comparison, we used the metrics described in subsection \ref{subsec:metrics}, trained the architectures with a maximum of 300 epochs on the training dataset and evaluated them on the evaluation dataset. We used the authors' codes where available and otherwise implemented the architecture ourselves. To limit computational costs, we used an early stopping mechanism for the training, that ended the training when the evaluation metrics did not improve over 10 epochs.

\begin{figure}[ht!]
\centering
\includegraphics[width=.9\textwidth]{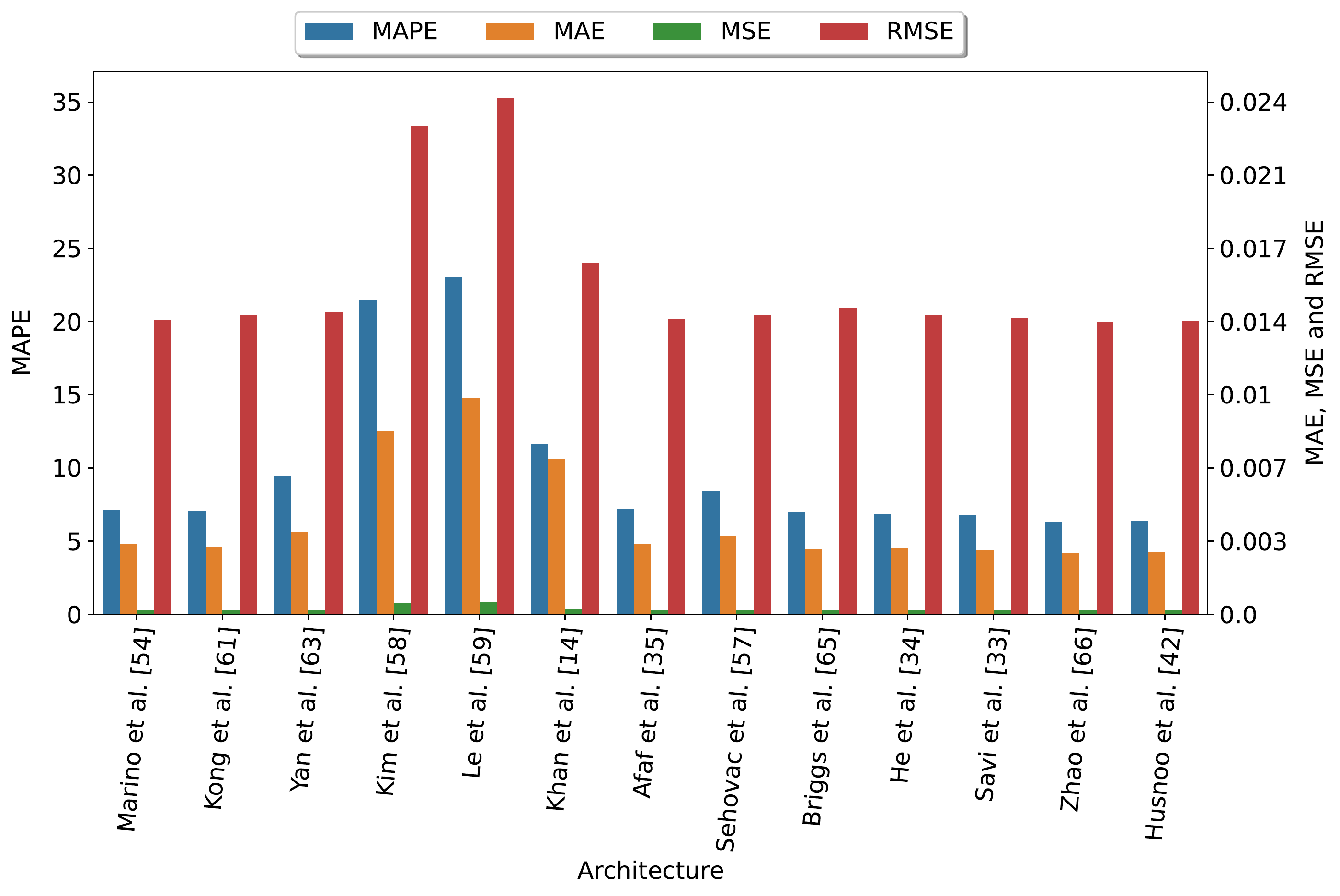}
\caption{RMSE, MSE, MAE, MAPE of the current literature applied to this paper's dataset. \label{fig:forecasting_literature_results}}
\end{figure}

In Figure~\ref{fig:forecasting_literature_results}, we illustrate the evaluation results for the twelve architectures we could replicate. Some architectures behaved worse on our dataset than on the dataset used by the respective authors. One possible reason for these differences could be scaling. Kim et al. ~\cite{kim2019predicting,app9204237}, for instance, worked with a non-scaled dataset. This means that depending on the standard deviation of the dataset $\sigma$, the error metrics can differ substantially. For instance, the MSE scales proportionally with the standard deviation: $MSE_{scaled} = MSE_{non-scaled} * \sigma$. To avoid this scaling effect, we calculated all metrics using standardized data (section \ref{sec:evaluation}).

Overall, the architectures in \cite{9469923,He2021ShortTermRL,Zhao2021ADP,Husnoo2022FedREPTH,8039509,en11113089,marino2016building} had  the lowest MAPE, from 6.7 to 7.1. From these, we selected Marino et al.'s  \cite{marino2016building} autoencoder architecture. Autoencoders are known to perform well even with non-idd data, so we selected the most performant autoencoder architecture among our shortlist of architecures. Marino et al.'s \cite{marino2016building} architecture uses a 50-neuron encoder layer, a 12-neuron latent space, a 50-neurons decoder layer, and two final layers with 100 and 1 neurons respectively. 

For our investigation of the effects of architectural complexity, we selected Khan et al.'s  \cite{s20051399} architecture as it performed best among the more complex architectures in our sample. Khan et al.'s \cite{s20051399} architecture is different from Marino et al.'s  \cite{marino2016building} in that it uses convolutional layers and LSTM.

\subsection{FL, differential privacy and secure aggregation set-up}
\label{subsec:securefederation}

For our simulations, we selected Fed-Avg over Fed-SGD as it requires fewer communication rounds and has better performance \cite{FEKRI2021107669, 8940936}. Moreover, we used a horizontal FL configuration as our clients represent different LCLIds but share the same feature space.

To implement DP, we followed the steps proposed by McMahan et al. \cite{mcmahan2018} rather than those of Chhachhi et al. \cite{dp_imperial_colleague} and Lu et al. ~\cite{lu2019blockchain}, in which noise is added to the dataset before the training. McMahan et al. \cite{mcmahan2018} propose the central server to add noise after aggregating the updates of the model weights at every training round (in Fed-Avg). In other words, it differs from canonical Fed-Avg, which aggregates model weights. 

The process proposed by McMahan et al. requires the definition of a query function sensitivity ($\mathbb{S}$) and a clipping strategy. The sensitivity of the query function determines the \textit{actuation range} of the added noise. It represents the Euclidean distance between two datasets ($C$) differing in at most one element $k$: $\mathbb{S}(\tilde{f}) = max_{C,k}\left \| \tilde{f}(C \cup \{k\} ) - \tilde{f}(c) \right \|_2 $ \cite{dwork2014algorithmic}. Considering McMahan et al.'s first lemma \cite{mcmahan2018} and assuming all clients are equally weighted, the sensitivity $\mathbb{S}$ is bounded as $\mathbb{S}(\tilde{f}(c)) \leq S/n $, with $n$ being the number of clients. The vectors in $\Delta_k$ include the different model updates computed among the clients. 

To bound the sensitivity of the query function, we needed to maintain the models' updates in a known range. One approach to ensure this range control is clipping model updates by a defined value before averaging. There are two strategies to clip the values of a neural network: 'per layer clipping', which applies clipping on a layer basis or 'flat clipping' which applies a clipping value to all the network parameters. Both clipping strategies project the values of the updates into a l2 sphere with the norm determined by the clipping value.

For both, per layer and flat clipping, there are two sub-strategies. One is to clip values using a fixed norm, known as fixed clipping. The second sub-strategy is called adaptive clipping \cite{andrew2021differentially}. It adapts the clipping norm based on a target quantile (i.e., 0.5) of the data distribution \cite{andrew2021differentially}.

For the sake of simplicity, we used flat clipping as $\Delta{}'_k = \pi(\Delta_k,S)$ with $S$ being the overall clipping value for the model updates. At the same time, we implemented both fixed and adaptive flat clipping strategies.

Once we had defined the query sensitivity and applied a flat clipping strategy, we evaluated how noise levels scale with the query sensitivity to obtain the minimum level of noise with a privacy guarantee. We added Gaussian noise as defined by: $N(0,{\sigma}^2)$ for $\sigma = z \cdot \mathbb{S}$, where $z$ is the noise scale and $\mathbb{S}$ is the sensitivity of the query.

The addition of noise determines the overall privacy protection ($\epsilon$) provided by DP. $\epsilon$ varies depending on the amount of noise added and the ratio of clients involved in the training ($Q$). $Q$ is the ratio of clients selected out of the total which will participate in the next round of training. More noise naturally means more privacy and a lower $\epsilon$. A higher $Q$, in turn, means less privacy and a higher $\epsilon$ \cite{abadi2016deep}.

To compute the privacy protection after a query, that is, each training round of our model, we used the privacy accountant provided by Renyi Differential Privacy (RDP)~\cite{Mironov_2017} as it provides a more detailed analysis of the privacy budget than the one created by \cite{mcmahan2018}. 

For SecAgg, we used the implementation provided by Bonawitz et al. \cite{SecAgg}. Their SecAgg implementation works as a plug-and-play algorithm that does not require any modification. 
We used SecAgg to ensure privacy-preserving communication between the central server and the clients. By using SecAgg in FL, clients can share their model weights without the central server or another client being able to reconstruct their weights ~\cite{shamir1979share}. 

\subsection{Model operation}
\label{subsec:model_operation}
\begin{figure}[ht!]
\centering
\includegraphics[width=\columnwidth]{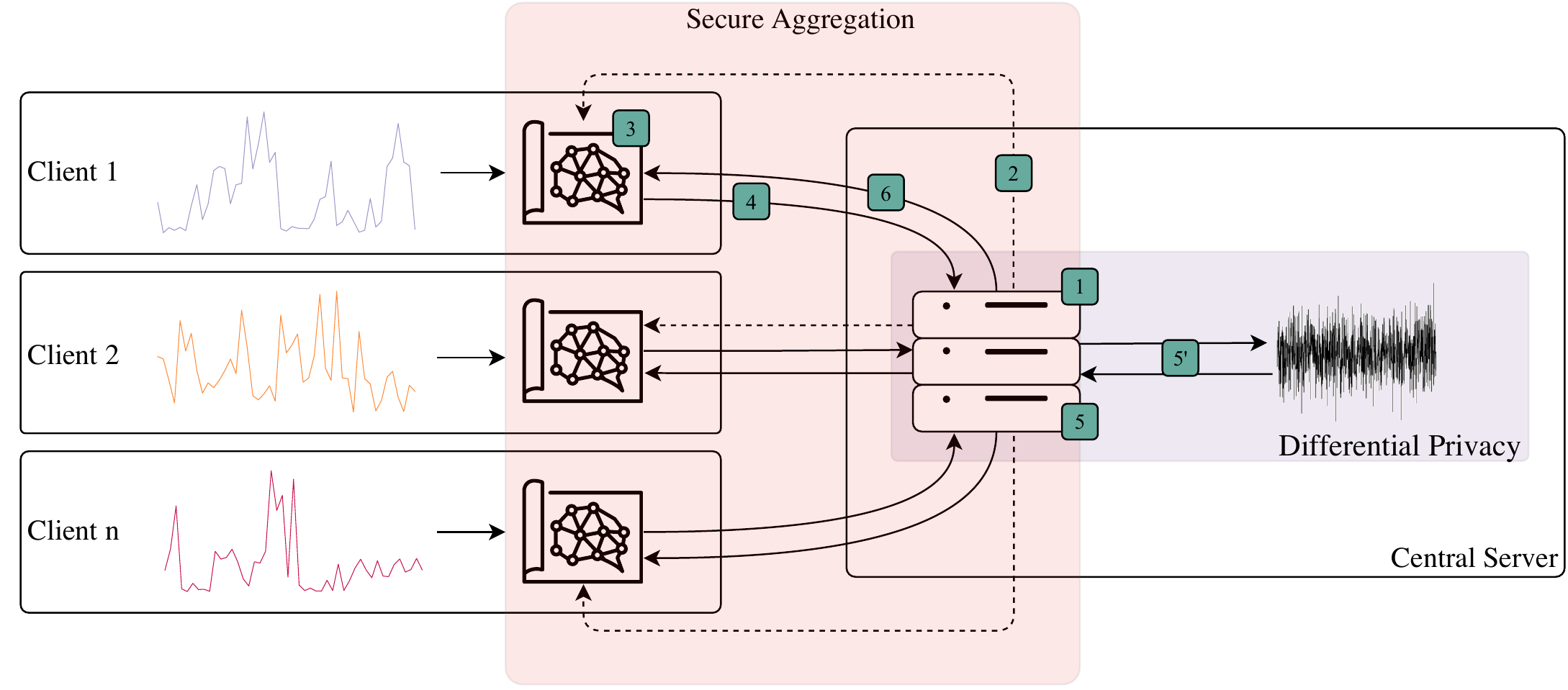}
\caption{Visual representation of our implementation of Federated Learning with privacy-preserving techniques.}
\label{FLdiagram}
\end{figure}

In this subsection, we describe how we trained the FL models. For this training, we used 6 steps. We illustrate these steps as well as the additional step that FL-DP requires in Figure ~\ref{FLdiagram}. FL-SecAgg requires a different additional step, namely the initial sharing of public keys between the clients and central server. Figure~\ref{FLdiagram} does not illustrate this additional public key sharing.

In step 1, the central server initializes the model using Glorot initialization \cite{glorot2010understanding}. In step two, the central server shares the model with the participating clients. In step three, a subset of clients are selected based on the ratio ($Q$). Each of these clients in this sub-set then trains the received model on its data. In step four, clients send their model updates to the central server. In step five, the central server averages the aggregated updates and adds noise drawn from a Gaussian distribution in the case of DP (5' in Figure~\ref{FLdiagram}). In step six, the central server returns the averaged updates to the clients. The central server and the clients repeated steps 2 to 6 until they reached 300 epochs.

\subsection{Scenario design}
\label{subsec:scenarios_desing}

Overall, we designed a set of six scenarios for our evaluation. Scenario 0 represents a hypothetical scenario in which all clients share their training data with the central server. This 'centralized setting' serves as a benchmark for the other scenarios. In Scenario A, we study the effects of moving from a centralized to a FL setting. In scenario B, we analyze the performance effect of clustering clients based on Pearson correlation. In scenario C, we evaluate the effect of a more complex NN architecture. Lastly, in Scenarios D and E, we study the effects of adding DP and SecAgg to the FL model. We summarize the specifications of the six scenarios in Table~\ref{table:scenarios}.

For scenarios 0, A, B, C and E, we ran eight simulations. These simulations evaluate the models’ performance with a growing number of clients (federation size). We used the following eight federation sizes: 2, 5, 8, 11, 14, 17, 20, and 23 clients. Each of these clients worked with data from one LCLid. We had to limit the maximum number of clients to 23 to control for computational cost as we simulated all clients and the communication between them in one virtual environment. In effect, every additional client did not add computational power but computational overhead. 

We provide an overview of the hyperparameters for scenarios 0, A, B, C and E in Table \ref{table:hyperparams}. Table \ref{tab:dp_implementation} provides the hyperparameters for the DP implementation in Scenario D.

\begin{table}[h!]
    \centering
    \caption{Scenarios considered.}
    \label{table:scenarios}
    \resizebox{\columnwidth}{!}{%
    \begin{tabular}{cccc}
        \toprule
        \textbf{Scenario} & \textbf{Privacy-Preserving Technique} & \textbf{NN Architecture} & \textbf{Imposed Correlation}  \\ \midrule
        
        \review{ \textbf{0}} & - & Marino et al. \cite{marino2016building} & \XSolidBrush  \\ 
        \textbf{A} & - & Marino et al. \cite{marino2016building} & \XSolidBrush  \\ 
        \textbf{B} & - & Marino et al. \cite{marino2016building} & \Checkmark \\ 
        \textbf{C} & - & Khan et al. \hspace{3.5pt} \cite{s20051399} & \XSolidBrush  \\ 
        \textbf{D} & Differential Privacy & He et al.\hspace{19pt}  \cite{He2021ShortTermRL} & \XSolidBrush  \\
        \textbf{E} & Secure Aggregation & Marino et al. \cite{marino2016building} & \XSolidBrush  \\ 
        \bottomrule
    \end{tabular}
    }
\end{table}

\begin{table}[h!]
    \centering
    \caption{\review{Hyperparameters for scenarios A,B,C and E. Those marked with * the ones used in scenario 0}.}
    \label{table:hyperparams}
    \resizebox{0.7\columnwidth}{!}{%
    \begin{tabular}{>{\raggedright}p{0.65\columnwidth}p{0.35\columnwidth}}
    \toprule
    \textbf{Parameter}                                  & \textbf{Value} \\
    \midrule
    Number of internal rounds before averaging          & 5              \\
    NN architecture                                     & Marino et al. \cite{marino2016building} * and Khan et al. \cite{s20051399} \\
    Ratio of clients involved per round (Q)             & 1              \\
    Total number of clients ($w$)                       & Subject to federation size \\
    Optimizer                                           & Adam *          \\
    Optimizer learning rate ($L_r$)                     & 0.01 *          \\
    Batch size                                          & 256  *          \\
    Number of communication rounds                      & 300  *          \\ 
    Number of internal epochs after training           & Not applicable           \\ 

    \bottomrule
 \end{tabular}
    }
\end{table}


\begin{table}[h!]
    \centering
    \caption{\review{Hyperparameters for scenario D}.}
    \label{tab:dp_implementation}
    \resizebox{0.7\columnwidth}{!}{%
    \begin{tabular}{>{\raggedright}p{0.65\columnwidth}p{0.35\columnwidth}}
    \toprule
    \textbf{Parameter}                                  & \textbf{Value} \\
    \midrule
    Number of internal rounds before averaging          & 5              \\
    NN Architecture                                      & He et al. \cite{He2021ShortTermRL}\\
    Ratio of clients involved per round ($Q$)           & 0.1              \\
    Total number of clients ($w$)                       & 100              \\
    Optimizer                                           & Adam           \\
    Optimizer learning rate ($L_r$)                     & 0.01           \\
    Batch size                                          & 64            \\
    Number of communication rounds                      & 100            \\ 
    Number of internal epochs after training            & 1           \\ 

    \bottomrule
 \end{tabular}
    }
\end{table}

\newpage

\section{Evaluation results}
\label{sec:evaluation}

\review{
\subsection{Scenario 0: Centralized setting}

Scenario 0 analyzes the performance of a centralized setting, in which the clients send their data to a central server that trains a single model on the aggregated data. Scenario 0 uses the NN architecture presented by Marino et al. \cite{marino2016building}. Similar to the architecture selection process, we employed an early stopper for Scenario 0 that terminated the training when there was no improvement in the validation metrics for more than 10 epochs.

In Table~\ref{table:scenario_0}, we collect the simulation results for scenario 0. The MSEs, RMSEs and MAEs are expressed in absolute values, the MAPEs in percentage points, and the average training time per epoch in second\,[s]

\begin{table}[h!]
    \centering
    \caption{Validation error metrics and computation time for one-hour-ahead prediction: Scenario 0.}
    \label{table:scenario_0} 
    \resizebox{.8\columnwidth}{!}{%
    \begin{tabular}{cccccc}
    \toprule
    \textbf{Central dataset size} & \textbf{MSE} & \textbf{RMSE} & \textbf{MAE} & \textbf{MAPE} & \textbf{Time per epoch\,[s]} \\
    \midrule
        \textbf{2}  & 0.00013 & 0.01158 & 0.00468 & 29.046 & 1.85  \\
        \textbf{5}  & 0.00012 & 0.01113 & 0.00308 & 9.068 & 6.01 \\
        \textbf{8}  & 0.00042 & 0.02067 & 0.00611 & 9.734 & 6.19 \\
        \textbf{11} & 0.00028 & 0.01681 & 0.00437 & 8.561 & 8.18 \\
        \textbf{14} & 0.00022 & 0.01514 & 0.00390 & 7.500 & 10.52 \\
        \textbf{17} & 0.00023 & 0.01519 & 0.00383 & 6.850 & 12.56 \\
        \textbf{20} & 0.00022 & 0.01498 & 0.00387 & 9.017 & 14.59 \\
        \textbf{23} & 0.00019 & 0.01388 & 0.00330 & 7.144 & 16.82 \\
    \bottomrule
    \end{tabular}
    }
\end{table}

Table~\ref{table:scenario_0} highlights that the overall performance of the centralized setting is very good, and that it remains almost constant for more than five clients with no evident variation in any of the metrics. The poor results in the two-client case could be the result of substantially different consumption patterns.

}
\subsection{Scenario A: standard federated learning setting}
\label{scenario_A}

We designed Scenario A to compare the 'centralized setting' in Scenario 0 with a FL setting, and to obtain a reference point for the other FL scenarios. Scenario A uses the NN architecture presented in \cite{marino2016building} and does not apply privacy-preserving techniques. Furthermore, we did not impose data correlation among the clients.

Table~\ref{table:scenario_1} presents the simulation results for Scenario A. The error metrics are expressed in absolute values and the average training time per epoch is expressed in seconds\,[s].

\begin{table}[h!]
    \centering
    \caption{\review{Validation error metrics and computation time for one-hour-ahead prediction: Scenario A.}}
    \label{table:scenario_1} 
    \resizebox{.8\columnwidth}{!}{%
    \begin{tabular}{cccccc}
    \toprule
    \textbf{Federation size} & \textbf{MSE} & \textbf{RMSE} & \textbf{MAE} & \textbf{MAPE} & \textbf{Time per round\,[s]} \\
    \midrule
        \textbf{2}  & 0.00015 & 0.01240 & 0.00516 & 30.1461 & 3.13  \\
        \textbf{5}  & 0.00022 & 0.01496 & 0.00468 & 16.2269 & 11.54 \\
        \textbf{8}  & 0.00058 & 0.02407 & 0.00745 & 11.9892 & 10.72 \\
        \textbf{11} & 0.00042 & 0.02049 & 0.00538 & 10.1082 & 13.39 \\
        \textbf{14} & 0.00035 & 0.01872 & 0.00542 & 10.1077 & 18.58 \\
        \textbf{17} & 0.00032 & 0.01787 & 0.00469 & 8.5392  & 21.05\\
        \textbf{20} & 0.00031 & 0.01775 & 0.00479 & 11.2933 & 25.10 \\
        \textbf{23} & 0.00028 & 0.01701 & 0.00478 & 10.8257 & 29.39 \\
    \bottomrule
    \end{tabular}
    }
\end{table}

Table~\ref{table:scenario_1} highlights that performance of FL models varies depending on the federation size. While MSEs, MAEs and RMSEs remain almost constant, there is a clear improvement in MAPEs. These results are in line with those by Savi et al. \cite{9469923} and Fekri et al. \cite{FEKRI2021107669} and indicate that larger federation sizes lead to more accurate FL models.

\begin{figure}[h!]
\centering
   \includegraphics[width=\textwidth]{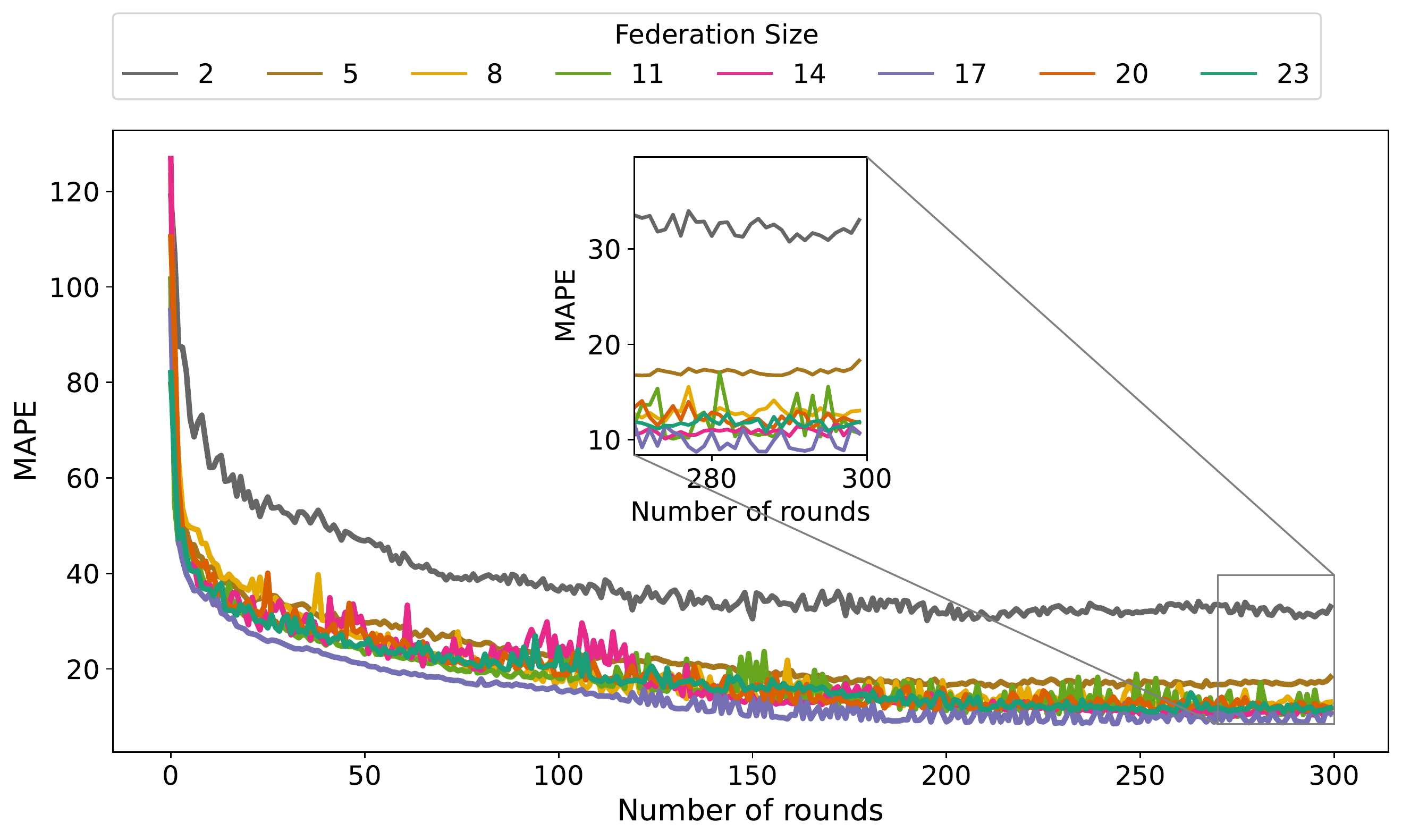}
\caption{Validation Mean Absolute Percentage Error (MAPE) per federation size in terms of training rounds for scenario A. \label{fig:scenario_1} }
\end{figure}

To better illustrate this effect, we plot how the MAPEs evolved for the eight federation sizes along the training rounds in Figure~\ref{fig:scenario_1}. Overall, we can observe a \textit{quasi-exponential} decrease over the 300 rounds, approaching final values between 6.8 and 29, which indicate reasonably good forecasts \cite{lewis1982industrial}. 

In comparison to Scenario 0, we can observe an average performance decrease between 20\% to 40\%. FL appears to perform significantly worse than a 'centralized' setting, which is in line with other comparable studies \cite{briggs2021federated, Husnoo2022FedREPTH, Lin2022PrivacyPreservingHC}.

Table~\ref{table:scenario_1} also highlights a trade-off between accuracy and computational time for federation size. As the number of clients increases, so does performance, but also computation time. This trade-off can present an important limitation for the use of FL.

\subsection{Scenario B: standard federated learning setting with imposed correlation}
\label{scenario_B}

In scenario B, we analyzed the performance of a standard FL setting with imposed correlation among the clients in the federation. We followed Lee et al. \cite{en13174408} and used Pearson correlation to identify and bundle clients (or LCLids) by correlated data. This way of bundling differs differs from the dominant k-means approach in prior literature and offers a more direct and simple view of the correlation between clients. More specifically, we pre-filtered our dataset for specific ACORNs (H and L). For these ACORNS, we then calculated all possible non-repeated combinations and calculated their correlations. For each federation size, we selected those combinations of clients with the highest correlations. 

We present the simulation results for Scenario B in Table~\ref{table:scenario_2}. The error metrics and the correlation rate are both expressed in absolute values. We omit the computation time because it was basically the same as in scenario A~(\ref{scenario_A}).

\begin{table}[ht!]
\centering
\caption{Validation error metrics and correlation rates for one-hour-ahead prediction: scenario B.}
\label{table:scenario_2} 
\resizebox{.8\columnwidth}{!}{%
\begin{tabular}{cccccc}
\toprule
\textbf{Federation size} & \textbf{MSE} & \textbf{RMSE} & \textbf{MAE} & \textbf{MAPE} & \textbf{Correlation rate} \\
\midrule

\textbf{2}  & 0.00002 & 0.00463 & 0.00170 & 4.54 & 0.62 \\
\textbf{5}  & 0.00015 & 0.01238 & 0.00373 & 9.77 & 0.51 \\
\textbf{8}  & 0.00022 & 0.01513 & 0.00426 & 8.91 & 0.49 \\
\textbf{11} & 0.00021 & 0.01465 & 0.00402 & 8.23 & 0.45 \\
\textbf{14} & 0.00020 & 0.01429 & 0.00390 & 8.66 & 0.42 \\
\textbf{17} & 0.00032 & 0.01805 & 0.00465 & 8.22 & 0.37 \\
\textbf{20} & 0.00029 & 0.01726 & 0.00428 & 8.38 & 0.34 \\
\textbf{23} & 0.00026 & 0.01640 & 0.00432 & 9.95 & 0.31 \\

\bottomrule
\end{tabular}
}

\end{table}

FL with imposed correlation performed better in almost every metric than FL without imposed correlation (Scenario A). The MSEs decreased by an average 35.87\%; RMSEs by 21.81\%; MAEs by 25.57\% and the MAPEs by 27.61\%. They nevertheless still trail Scenario 0 by 6.35\% on average. Moreover, these values are subject to some caveats. Our model with two clients had a correlation rate of 0.62, which led to a 75\% better performance than the two-client case in Scenario A. Moreover, the performance of the model with 17 clients was worse than the same model in Scenario A, and 45\% of the error metrics in Scenario B were better than those in scenario 0.

These results align well with similar studies, such as \cite{FEKRI2021107669, Biswal2021AMIFMLAP, He2021ShortTermRL} or \cite{9469923}, where the application of k-means to cluster customers leads to performance improvements between 10\% and 15\%. 

Overall, scenario B suggests that clustering based on Pearson correlation among the clients in a federation can substantially improve the performance of FL-based STLF. Specifically, utilities, energy providers, and DSOs could leverage simple socio-economic factors (ACORNS) and historical, individual smart meter data to cluster their residential customers into correlated groups. Each cluster can use a different FL model to reduce imbalance costs for inaccurate forecasts and offer tailored demand-side management programs.

\subsection{Scenario C: standard federated learning setting with a more complex neural network architecture}
\label{scenario_C}

In scenario C, we explore how a more complex NN architecture (\cite{s20051399}) impacts the performance of FL-based STLF. The motivation for scenario C is rooted in the trend to use ever more complex machine learning architectures in the hope of catching patterns invisible to less complex architectures. At the same time, it is unclear whether larger architectures increase performance. 

To account for the size of the model in \cite{s20051399} and its computational burden, we implemented three modifications to the set-up of our simulation environment. The first modification concerns the GPUs. For each of the Nvidia Tesla allocated on the HPC, we created two virtual cards, resulting in four cards we could use for our simulation. The second modification is related to the batch size, which we increased from 100 to 200. Increasing the batch size can help to prevent or limit overfitting since there are more data entries available to compute the loss of the model. Finally, we modified the model in \cite{s20051399} by transforming the initially proposed LSTM layers to CuDNNLSTM~\cite{appleyard2016optimizing}. The transformation enabled the LSTMs to use the Compute Unified Device Architecture (CUDA) kernel of our Tesla GPUs.

The simulation results of scenario C are presented in Table~\ref{table:scenario_3}. The results clearly indicate the increased computational costs of training a FL model with a complex architecture. The computational time is almost twice as high as in scenarios A and B. On the other hand, the performance of the model with the more complex architecture was worse that of the smaller model's for all federation sizes and all metrics, ranging from 50\% up to 142\%.

\begin{table}[ht!]
    \centering
    \caption{Validation error metrics and computation time for one-hour-ahead prediction: scenario C.}
    \label{table:scenario_3} 
    \resizebox{.8\columnwidth}{!}{%
    \begin{tabular}{cccccc}
        \toprule
        \textbf{Federation size} & \textbf{MSE} & \textbf{RMSE} & \textbf{MAE} & \textbf{MAPE} & \textbf{Time per round\,[s]} \\ \midrule
        \textbf{2}  & 0.00024 & 0.01550 & 0.00720 & 31.50674 & 6.25  \\
        \textbf{5}  & 0.00052 & 0.02289 & 0.01282 & 33.42653 & 21.10 \\
        \textbf{8}  & 0.00117 & 0.03433 & 0.01754 & 20.92209 & 20.43 \\
        \textbf{11} & 0.00115 & 0.03398 & 0.01495 & 21.93438 & 30.34 \\
        \textbf{14} & 0.00087 & 0.02955 & 0.01404 & 18.44877 & 34.52 \\
        \textbf{17} & 0.00077 & 0.02783 & 0.01080 & 13.80498 & 40.59 \\
        \textbf{20} & 0.00081 & 0.02858 & 0.01435 & 24.28874 & 50.19 \\
        \textbf{23} & 0.00061 & 0.02486 & 0.01059 & 19.02717 & 59.76 \\
        \bottomrule
    \end{tabular}
    }
\end{table}

These results suggest a clear case of overfitting. Overfitting is generally defined as the lack of generalization of a model. An overfitted model crosses the line between learning tendencies or patterns and \textit{memorizing} the data received as input. 

\begin{figure}[ht!]
\centering
\includegraphics[width=\columnwidth]{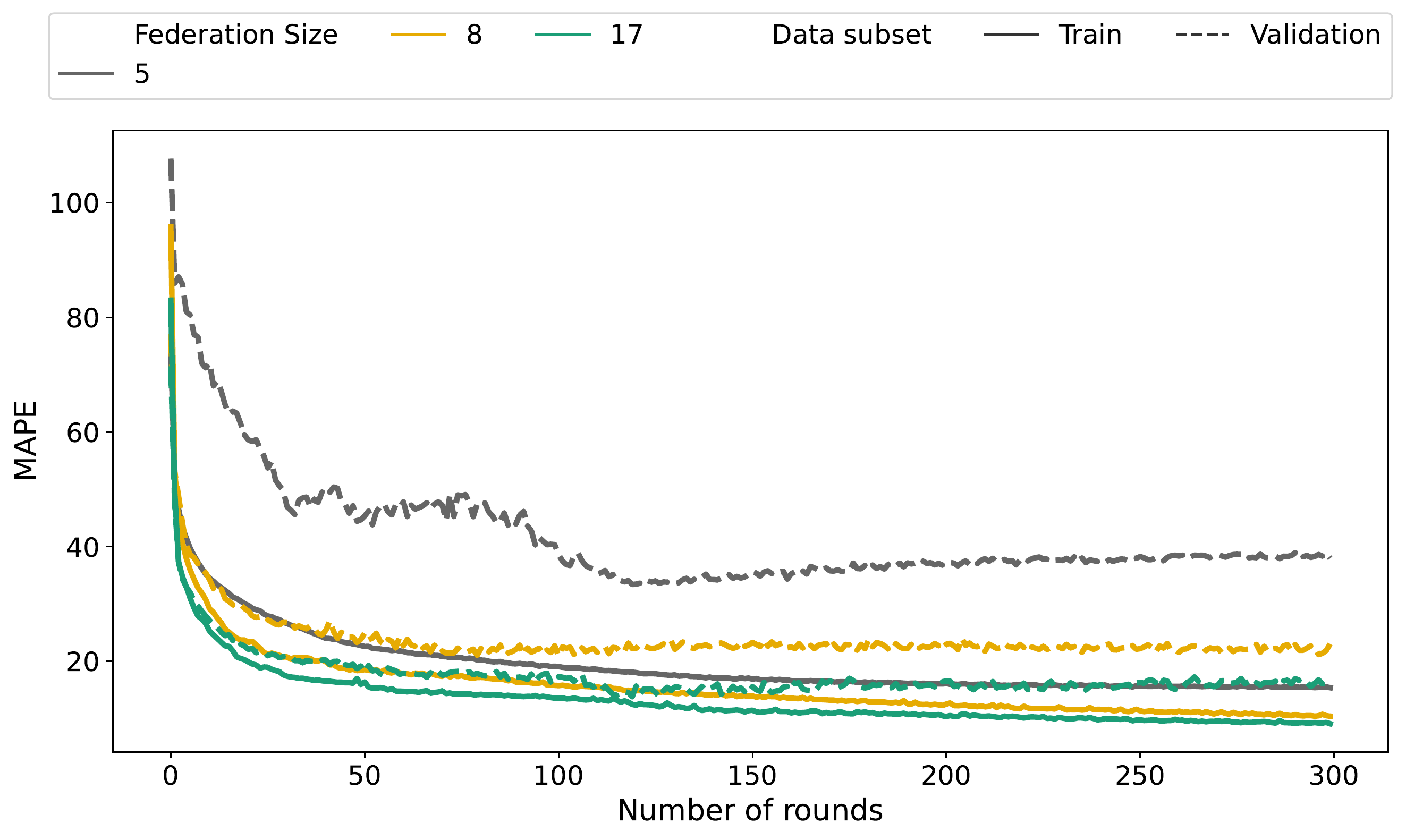}
\caption{Validation and Training MAPEs for federation sizes 5,8, and 17 in Scenario C. \label{fig:results_scenarioC}}
\end{figure}

Figure~\ref{fig:results_scenarioC} provides a visualization of this overfitting. The performance on the training subset is represented by the solid lines, while the performance on the validation subset is visualized by the dotted lines. The dotted lines begin to increase again after round 120, whereas the solid lines decrease as the model is over-fitted to the training data. .

In effect, scenario C offers a cautionary tale for utilities, energy providers, and DSOs that want to use FL for short-term load forecasting. Not only are more complex FL architectures more expensive and detrimental to the environment \cite{hao2019training}, they are also more sensitive to handle.

\subsection{Scenario D: privacy-preserving federated learning setting with differential privacy}
\label{scenario_D}

Scenario D focuses on adding DP to FL and how this impacts the performance of FL-based STLF. Furthermore, we compare two \textit{flat clipping} approaches: fixed and adaptive clipping, as described in subsection~\ref{subsec:securefederation}. 

In scenarios A and B, we used Marino et al.'s model \cite{marino2016building} as the baseline architecture. Encoder-decoder architectures can cope well with outliers due to their capacity to abstract information into the latent space. This capacity is very beneficial for FL where different clients can have substantially different data points. However, we found that these architectures are substantially more vulnerable to noise than standard stacked LSTM networks. One reason for this vulnerability could be that they compact information from a higher dimensional space into a smaller one. Adding noise to the weights of this latent space will have a multiplicative effect on the model's output in the decoder phase. To avoid such encoder-decoder noise problems for our DP simulation, we changed the architecture in Scenario D to a two-layer LSTM with 50 neurons each, and a final dense layer as in He et al. \cite{He2021ShortTermRL}. 

DP offers two approaches to obtain a high privacy budget given a defined amount of noise: reduce the ratio of clients that participate in each training round ($Q$), retrain the model locally for several epochs on client data, find a lower $\delta$, and/or increase the noise scale ($z$). For Scenario D, we employed a ratio of $Q=0.1$. With $Q=0.1$, a total of 100 clients and without the addition of privacy preserving techniques, our model had a MAE of 0.00300, a MSE of 0.012, a RMSE of 0.01114, and a MAPE of 8.3846, which matches results in Scenario A. 

Moreover, we considered recommendations by Zhao et al. \cite{Zhao2021ADP} and Xu et al. \cite{Xu2021LSTMSR} to introduce local re-training. Specifically, they propose to conduct several local training rounds on each client between each aggregation with DP to better fit the local models. Yet, we found that these repeated rounds didn't improve performance so we chose to use just one local training round. However, we did optimize the $\delta$ to $\delta = 4e^{-3}$ as proposed by Zhao et al. \cite{Zhao2021ADP}.

The first strategy we implemented was fixed clipping following the two main steps in McMahan et al. \cite{mcmahan2018}. In the first step, we determined the lowest possible clipping value ($S$) as being too low clipping values can negatively affect the convergence rate as they clip all values bigger than $S$. We treated  $S$ as a hyper-parameter and  used an iterative approach to find the lowest possible clipping value. Specifically, we followed McMahan et al. \cite{mcmahan2018} and used iterative steps of 0.1 for $S$, starting with $S=0.1$ until $S=0.7$. We present the error metrics for the different $S$ values in Table~\ref{table:comparison_s}.~\footnote{Setting a fixed value for the clipping slows the training process significantly. The values in Table~\ref{table:comparison_s} are the validation metrics after 2000 communication rounds. Without any clipping strategy, the models converge at an earlier rate (see figure~\ref{fig:scenario_1})}. 

Based on these iterations, we selected $S \approx 0.3$ as our fixed clipping value. It is the lowest clipping value with comparatively good error metrics and the marginal increase in error metrics from lowering $S$ increases disproportionately below $\approx 0.3$.

\begin{table}[ht!]
    \centering
    \caption{Validation error metrics for different clipping values for one-hour-ahead prediction with the sample client ratio $Q=0.1$ and total number of clients $w=100$: scenario D. } 
    \label{table:comparison_s} 
     \resizebox{.6\columnwidth}{!}{%
    \begin{tabular}{cccccc}
    \toprule
    \textbf{S} & \textbf{MSE} & \textbf{RMSE} & \textbf{MAE} & \textbf{MAPE} \\ 
    \midrule
    
        0.10 & 0.00043 & 0.02094 & 0.00628 & 10.69357 \\
        0.20 & 0.00035 & 0.01884 & 0.00502 & 8.89023  \\
        0.30 & 0.00038 & 0.01969 & 0.00496 & 8.00244 \\
        0.40 & 0.00038 & 0.01963 & 0.00486 & 7.71642   \\
        0.50 & 0.00039 & 0.01978 & 0.00493 & 7.92688  \\
        0.60 & 0.00034 & 0.01869 & 0.00477 & 7.81763  \\
        0.70 & 0.00036 & 0.01915 & 0.00484 & 7.53057  \\
    \bottomrule
    \end{tabular}
}
\end{table}

Once we had identified the lowest possible clipping value $S$, the second step was to identify a tolerable level of noise. With $S=0.3$, a total number of clients $w = 100$, and $Q=0.1$, we applied $\mathbb{S} = \nicefrac{S}{Qw}$ to calculate the standard deviation of the noise level $\sigma = z \cdot \mathbb{S} $. Similarly with the approach that we took with $S$, we treated $z$ as a hyper-parameter and ranged it from \review{0.1 to 0.9}

\begin{table}[ht]
    \centering
    \caption{\review{Exploration of the different noise levels, in bold the hyper-parameter  $\mathbf{z}$ that defines the amount of noise.}}
    \label{table:S-values} 
    \resizebox{.5\columnwidth}{!}{%
    \begin{tabular}{ccccc}
    \toprule
    $\mathbf{Qw}$ & $\mathbf{S}$ & $\pmb{\mathbb{S}} = \nicefrac{\mathbf{S}}{\mathbf{Qw}}$ & $\mathbf{z}$ & $\pmb{\sigma} = \mathbf{z} \cdot \pmb{\mathbb{S}} $ \\
    \midrule
    10   & 0.3  & 0.03    & 0.1   & 0.003   \\
    10   & 0.3  & 0.03   & 0.2    & 0.006   \\
    10   & 0.3  & 0.03   & 0.3    & 0.009   \\
    10   & 0.3  & 0.03   & 0.4    & 0.012   \\
    10   & 0.3  & 0.03   & 0.5    & 0.015   \\
    10   & 0.3  & 0.03   & 0.6    & 0.018   \\
    10   & 0.3  & 0.03   & 0.7    & 0.021   \\
    10   & 0.3  & 0.03   & 0.8    & 0.024   \\
    10   & 0.3  & 0.03   & 0.9    & 0.027   \\

    \bottomrule    
    \end{tabular}
}
\end{table}

In Table~\ref{table:S-values}, we present the performance metrics for each of the $z$ variations. Each of the explored $z$ values represents a different level of noise added to the federated model. Intuitively, there is a trade-off between the amount of noise and performance, whereby more noise (increase in $z$) reduces performance. This trade-off dynamic is clear from the error metrics in Table~\ref{table:scenario_3_fixedS}. Nevertheless, the overall error metrics for DP based on fixed clipping are generally low and indicate good forecasting performance. 

Concurrently, more noise also means better privacy, as indicated by the increasing privacy guarantees in column three of Table~\ref{table:scenario_3_fixedS}. We calculated these guarantees using the Rényi Differential Privacy Accountant \cite{Mironov_2017}. The highest amount of noise we examined ($z$=0.9) provides a privacy guarantee of (4.2, $4e^{-3}$), which is close to perfect privacy ($\epsilon = 0$). In effect, scenario D demonstrates that adding DP to FL maintains comparatively good performance and offers high privacy guarantees.

\begin{table}[h!]
    \centering
    \caption{\review{Validation error metrics with $S=0.3$ and a varying noise scale $\mathbf{z}$ from \review{0.1 to 0.9} for one hour-ahead-prediction with the sample client ratio $Q=0.1$ and total number of clients $w=100$ after one epoch of local training.}}
    \label{table:scenario_3_fixedS} 
    \resizebox{\textwidth}{!}{%
    \begin{tabular}{ccccccc}
        \toprule
        \textbf{Noise scale ($\mathbf{z}$)} & \textbf{Privacy Guarantee (\pmb{$\epsilon,\delta$})} & \textbf{MSE} & \textbf{RMSE} & \textbf{MAE} & \textbf{MAPE} & \textbf{Timer per round [s]} \\
        \midrule
        0.1 &   (911,$4e^{-3}$)      & 0.00010 & 0.00946 & 0.00272 & 7.5426     & 86.74   \\
        0.2 &   (190,$4e^{-3}$)      & 0.00010 & 0.00957 & 0.00312 & 8.8930     & 85.11   \\
        0.3 &   (69.3,$4e^{-3}$)     & 0.00010 & 0.00959 & 0.00309 & 8.4391     & 87.48   \\
        0.4 &   (32.4,$4e^{-3}$)     & 0.00010 & 0.00962 & 0.00321 & 9.1156     & 84.66   \\
        0.5 &   (17.9,$4e^{-3}$)     & 0.00011 & 0.00971 & 0.00340 & 9.7164     & 88.52   \\
        0.6 &   (11.2,$4e^{-3}$)     & 0.00011 & 0.00972 & 0.00344 & 9.9693     & 84.28   \\
        0.7 &   (7.58,$4e^{-3}$)     & 0.00011 & 0.00979 & 0.00354 & 10.0378    & 81.46   \\
        0.8 &   (5.5,$4e^{-3}$)      & 0.00013 & 0.01075 & 0.00519 & 15.6755    & 82.08   \\
        0.9 &   (4.2,$4e^{-3}$)      & 0.00011 & 0.00991 & 0.00372 & 10.6031    & 87.48   \\

        \bottomrule

    \end{tabular}
    }
\end{table}


The second clipping strategy that we analyzed is adaptive clipping. With adaptive clipping, clipping value are calculated automatically. To evaluate this approach, we used Andrew et al.'s adaptive clipping implementation \cite{andrew2021differentially}, in which the algorithm iteratively (per communication round) adjusts the norm clip, trying to approximate it to a predefined quantile (0.5 in our case). 

This data quantile approximation expends privacy budget as it queries the data. To prevent this \textit{privacy leakage} Andrew et al.~\cite{andrew2021differentially} propose to add noise during the approximation. This noise ($\sigma_b$) is defined by 0.05 times the number of clients per round, in our case $\sigma_b = 0.5$. This addition of noise has a slight affect on the total privacy guarantee of the model. It results in increased effective noise as $z_{\Delta} = (z^{-2} - (2\sigma_b)^{-2})^{-1/2} $.

\begin{figure}[h!]
    \centering
    \includegraphics[width=\columnwidth]{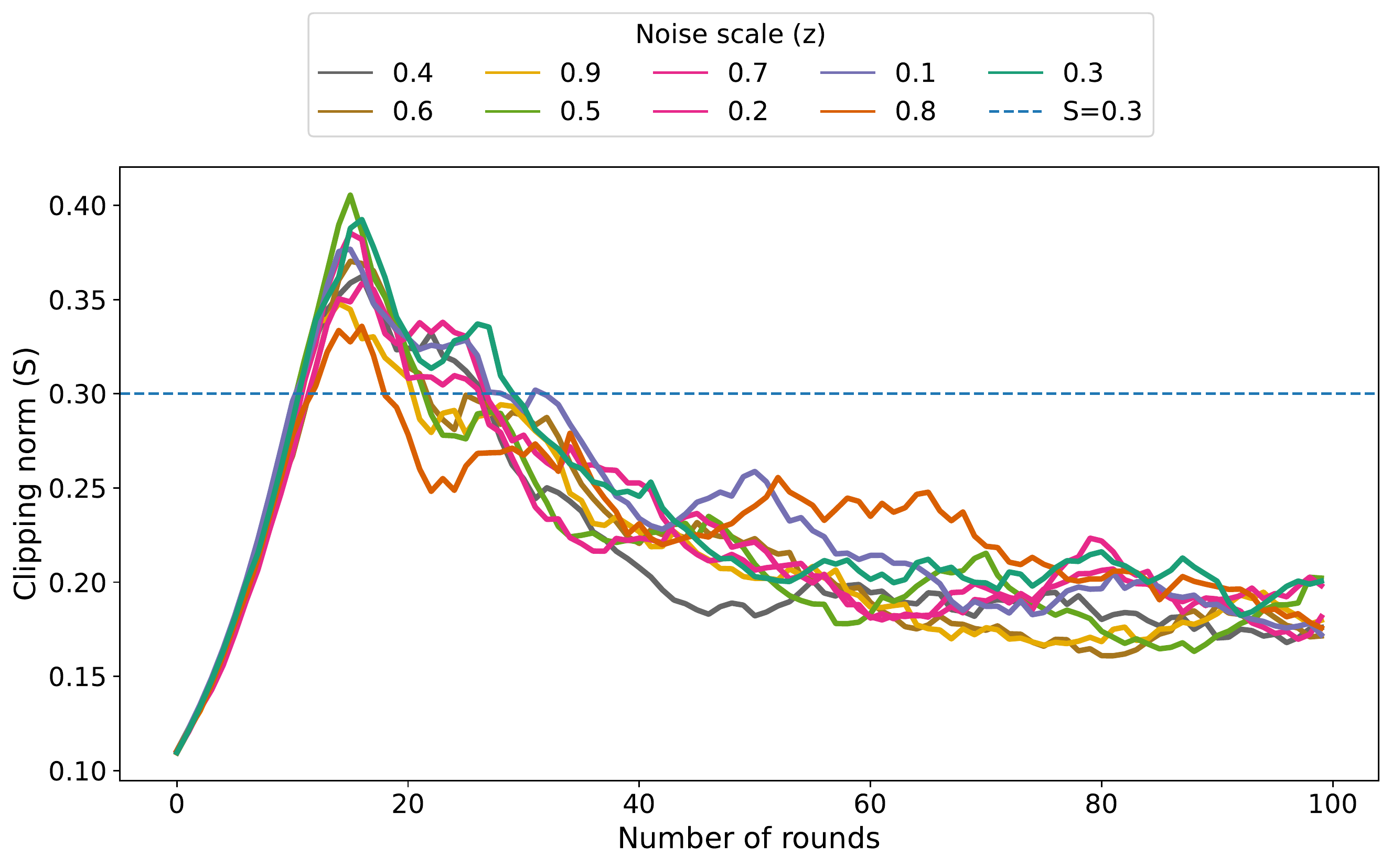}
    \caption{Evolution of the adaptive clipping norm at different noise levels $z$ (0.1, 0.2, 0.3, 0.4, 0.5, 0.6, 0.7, 0.8, 0.9) using as initial clipping value $C^0=0.1$ and the step factor for the geometric updates $\eta C=0.2$.}
    \label{fig:normclip}
\end{figure}

Figure~\ref{fig:normclip} highlights the adaptive adjustments of the clipping value over the training rounds. There is a sharp increase in the clipping norm at the beginning of the training rounds due to the low initial clipping value $C^0=0.1$. Such a low quantile allows only a few data points to participate in selecting the clipping value. The smaller the quantile, the fewer data points participate and thus, it is more difficult to estimate the optimal clipping value. 

As in our case, the adaptive clipping algorithm may overshoot as a result and increase the clipping norm to higher values. After this overshot, the adaptive clipping algorithm correctly approximates the optimal clipping value $S \approx 0.2$.

We present the simulation results for adaptive clipping in Table~\ref{table:scenario_DP_adaptive}. On average, adaptive clipping outperformed fixed clipping by 9\%. Moreover, the privacy guarantee is close to perfect privacy (3.9,$4e^{-3}$)

Adaptive clipping appears not only more attractive from a performance and privacy perspective. It is also easier to use in terms of performance and privacy. Fixed clipping requires an initial and computationally expensive manual step to identify an appropriate clipping value, whereas, in adaptive clipping, this value is calculated automatically in the training rounds. Thus, DP with adaptive clipping presents the more convenient choice for residential STLF.

\begin{table}[!h]
    \centering
    \caption{\review{Validation error metrics with adaptive clipping at different noise levels from 0.1 to 0.9 using as initial clipping value $C^0=0.1$ and the step factor for the geometric updates $\eta C=0.2$ for one hour ahead prediction with the sample client ratio $Q=0.1$ and total number of clients $w=100$ after one epoch of local training.}}
    \label{table:scenario_DP_adaptive} 
    \resizebox{\textwidth}{!}{%
    \begin{tabular}{cccccccc}
        \toprule
        \textbf{Noise scale ($\mathbf{z}$)} &
        \textbf{Effective noise (\pmb{$z_{\Delta}$})} & \textbf{Privacy Guarantee (\pmb{$\epsilon,\delta$})} & \textbf{MSE} & \textbf{RMSE} & \textbf{MAE} & \textbf{MAPE} & \textbf{Time per round [s]} \\
        \midrule
        0.1     & 0.100  & (910.0,$4e^{-3}$)    &  0.00010 & 0.00936  & 0.00276     & 7.9966 & 84.39   \\
        0.2     & 0.200  & (189.4,$4e^{-3}$)    & 0.00010 & 0.00930  & 0.00260      & 7.3866 & 88.41   \\
        0.3     & 0.300  & (68.7,$4e^{-3}$)     & 0.00009 & 0.00930  & 0.00257      & 7.0985 & 85.30   \\
        0.4     & 0.402  & (31.9,$4e^{-3}$)     & 0.00010 & 0.00945  & 0.00292      & 8.2810 & 86.92   \\
        0.5     & 0.504  & (17.5,$4e^{-3}$)     & 0.00010 & 0.00948  & 0.00301      & 9.0461 & 88.57   \\
        0.6     & 0.607  & (10.8,$4e^{-3}$)     & 0.00010 & 0.00955  & 0.00302      & 8.8343 & 86.27   \\
        0.7     & 0.711  & (7.2,$4e^{-3}$)      & 0.00010 & 0.00961  & 0.00317      & 9.4312 & 87.68   \\
        0.8     & 0.817  & (5.2,$4e^{-3}$)      & 0.00010 & 0.00955  & 0.00325      & 9.6126 & 88.27   \\
        0.9     & 0.924  & (3.9,$4e^{-3}$)      & 0.00010 & 0.00955  & 0.00319      & 9.2953 & 87.93   \\

        \bottomrule
    \end{tabular}
    }    

\end{table}

\begin{figure}[ht!]
\centering
\includegraphics[width=0.9\columnwidth]{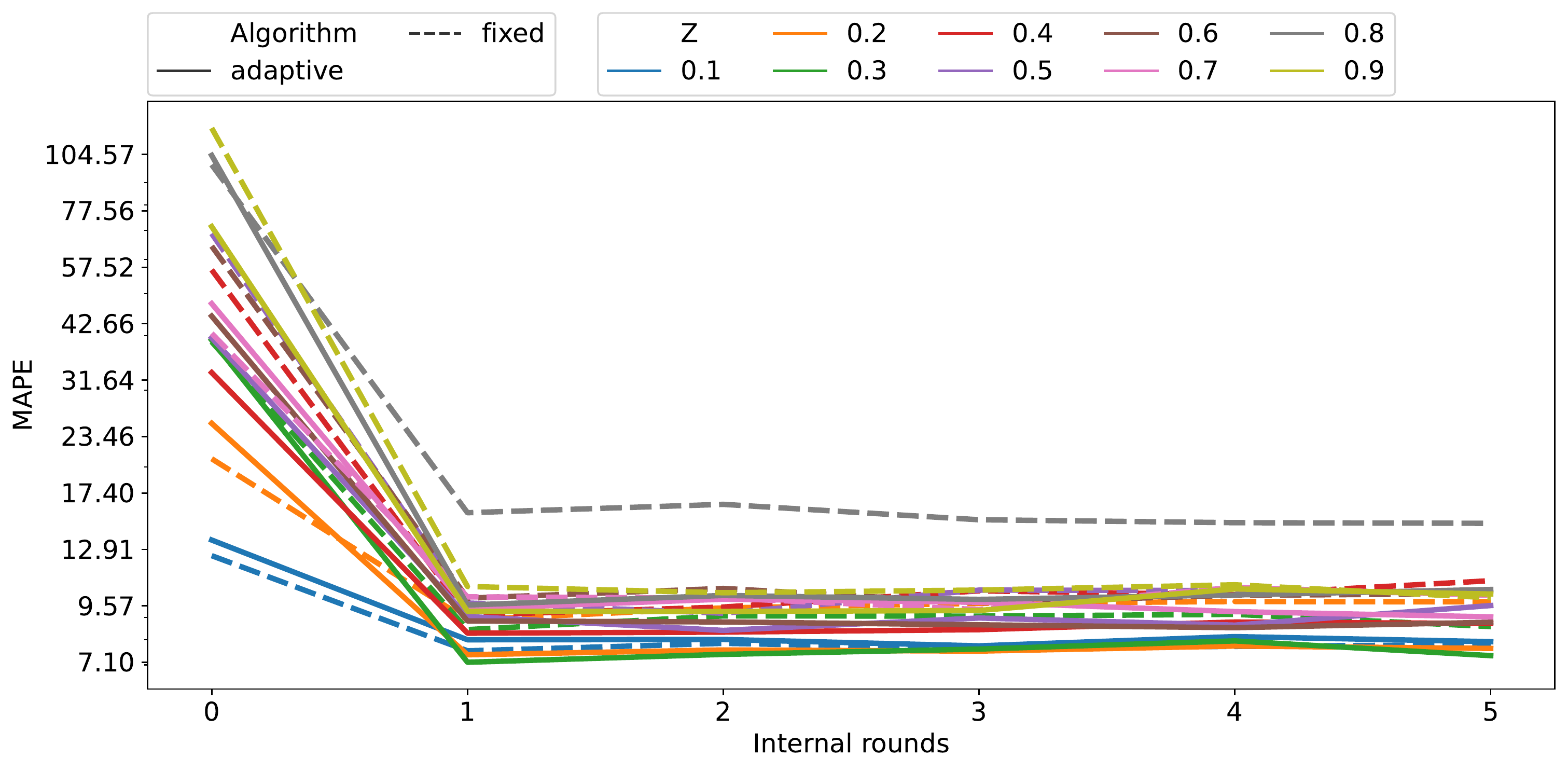}
\caption{Validation Mean Absolute Percentage Error (MAPE) per local training epoch for adaptive and fixed DP. \label{fig:local_training}}
\end{figure}

The results we present in Tables \ref{table:scenario_3_fixedS} and \ref{table:scenario_DP_adaptive} are those after the local training round suggested by Zhao et al. \cite{Zhao2021ADP}. Unlike Zhao et al. \cite{Zhao2021ADP}, who worked with five local training round, we used only one as additional rounds did not significantly improve performance (Figure~\ref{fig:local_training}). Nevertheless, clients profited from local training with negligible computational overhead.

\subsection{Scenario E: privacy-preserving federated learning setting with secure aggregation}
\label{scenario_E}

In this scenario, we examine SecAgg as an alternative technique to add privacy to FL. Whereas DP adds random noise to model updates, SecAgg targets the communication and aggregation of the clients' model updates. Hence, there is no trade-off as in scenario D, where it is important to find an adequate noise level. 

\begin{figure}[h!]
    \centering
    \includegraphics[width=.9\textwidth]{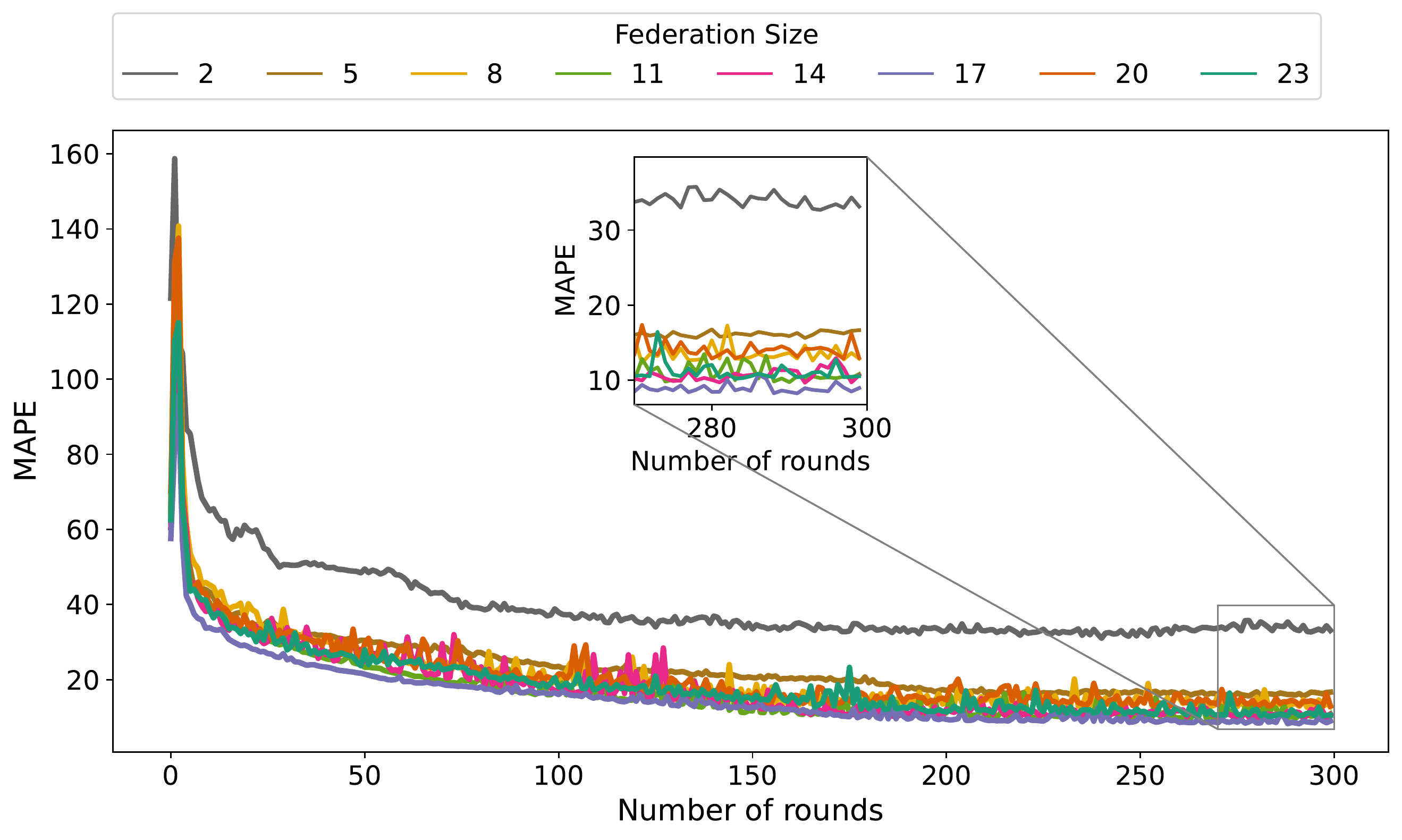}
    \caption{Validation Mean Absolute Percentage Error (MAPE) per LCLids federation size in terms of training rounds for Scenario E. \label{fig:results_scenarioE}}
\end{figure}

Similar to scenarios A, B and C, we present the simulation results for the eight federation sizes in Table~\ref{table:scenario_E}. We express the error metrics in absolute values and the average computation time in seconds\,[s]. Furthermore, we complement the results with Figure~\ref{fig:results_scenarioE}. It depicts the MAPE, following a similar curve as in Scenario A.

Table~\ref{table:scenario_E} shows that the use of SecAgg affects computation time only marginally. As SecAgg does not add any noise, it also provides less burden than DP. Consequently, SecAgg presents a more performant alternative for residential STLF with the cost of an extra 30\% of computation time. However, it is important to note that SecAgg does not provide complete privacy because latent patterns could still point toward the original data subject. More specifically, Model Inversion (MI) attacks could reconstruct the original training data from the model parameters \cite{fredrikson2015model}.

\begin{table}[ht!]
    \centering
    \caption{Error metrics and computation time for one-hour-ahead prediction using SecAgg: scenario E on test set.}
    \label{table:scenario_E} 
    \resizebox{.8\columnwidth}{!}{    
    \begin{tabular}{cccccc}
    \toprule
    \textbf{Federation size} & \textbf{MSE} & \textbf{RMSE} & \textbf{MAE} & \textbf{MAPE} & \textbf{Time per round\,[s]}\\ \midrule
        \textbf{2}  & 0.00017 & 0.01324 & 0.00532 & 31.01177 & 4.54  \\
        \textbf{5}  & 0.00018 & 0.01348 & 0.00431 & 15.60893 & 13.23 \\
        \textbf{8}  & 0.00060 & 0.02457 & 0.00759 & 12.28532 & 13.34 \\
        \textbf{11} & 0.00039 & 0.01996 & 0.00523 & 9.65965  & 18.21 \\
        \textbf{14} & 0.00034 & 0.01864 & 0.00503 & 9.67057  & 22.25 \\
        \textbf{17} & 0.00033 & 0.01820 & 0.00466 & 8.25973  & 26.70 \\
        \textbf{20} & 0.00033 & 0.01836 & 0.00522 & 12.88359 & 34.64 \\
        \textbf{23} & 0.00028 & 0.01683 & 0.00453 & 10.19247 & 38.10  \\
    \bottomrule
     
    \end{tabular}
    }
\end{table}

\newpage
\subsection{\review{Comparison across the scenarios}}

\begin{figure}[h!]
    \centering
    \includegraphics[width=.6\textwidth]{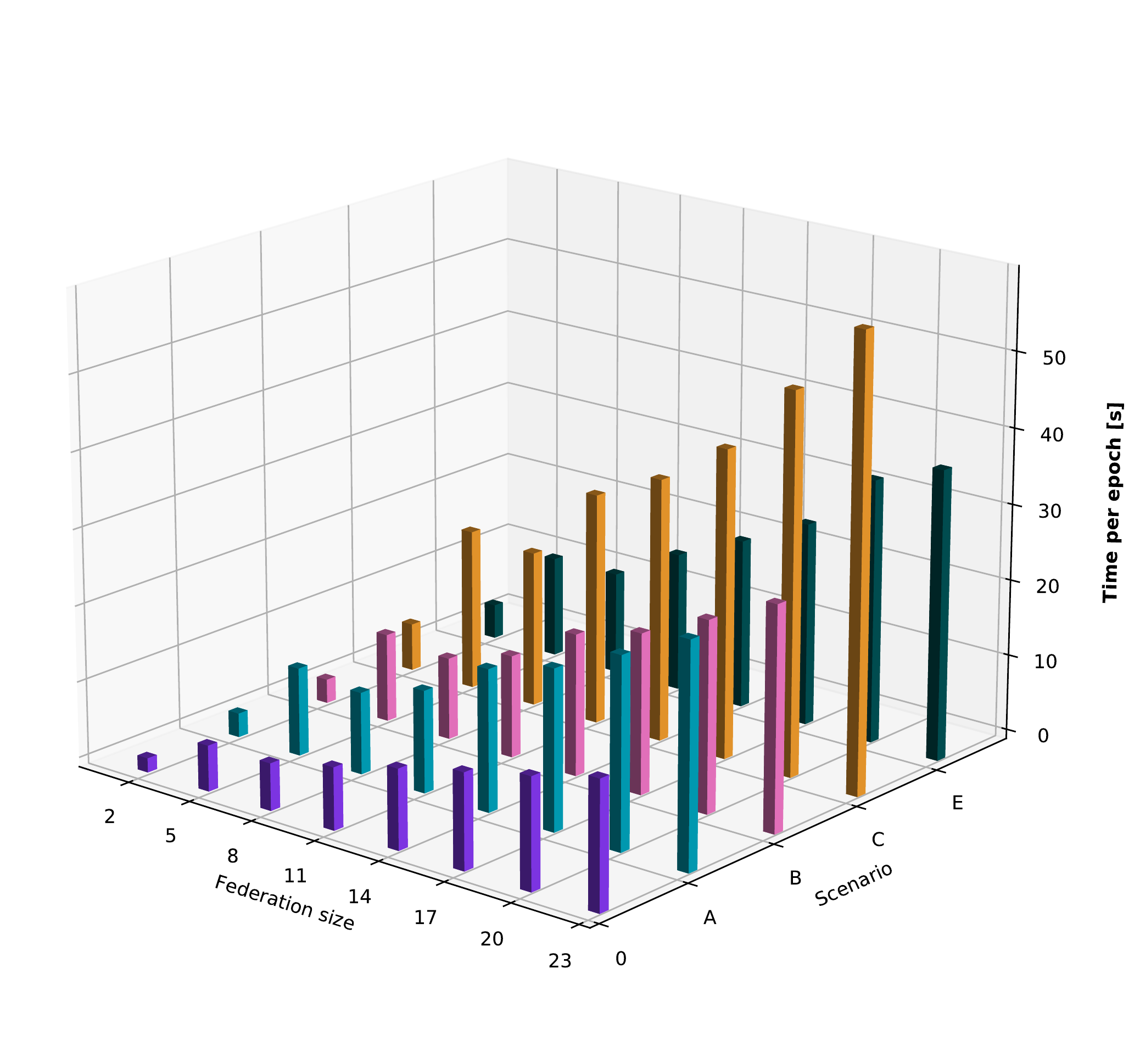}
    \caption{Comparison of computation time across Scenarios 0, A, B, C, and E.}
    \label{fig:computation_time_comparison}
\end{figure}

\begin{figure*}[h!]
        \centering
        \begin{subfigure}[b]{0.475\textwidth}
            \centering
            \includegraphics[width=\textwidth]{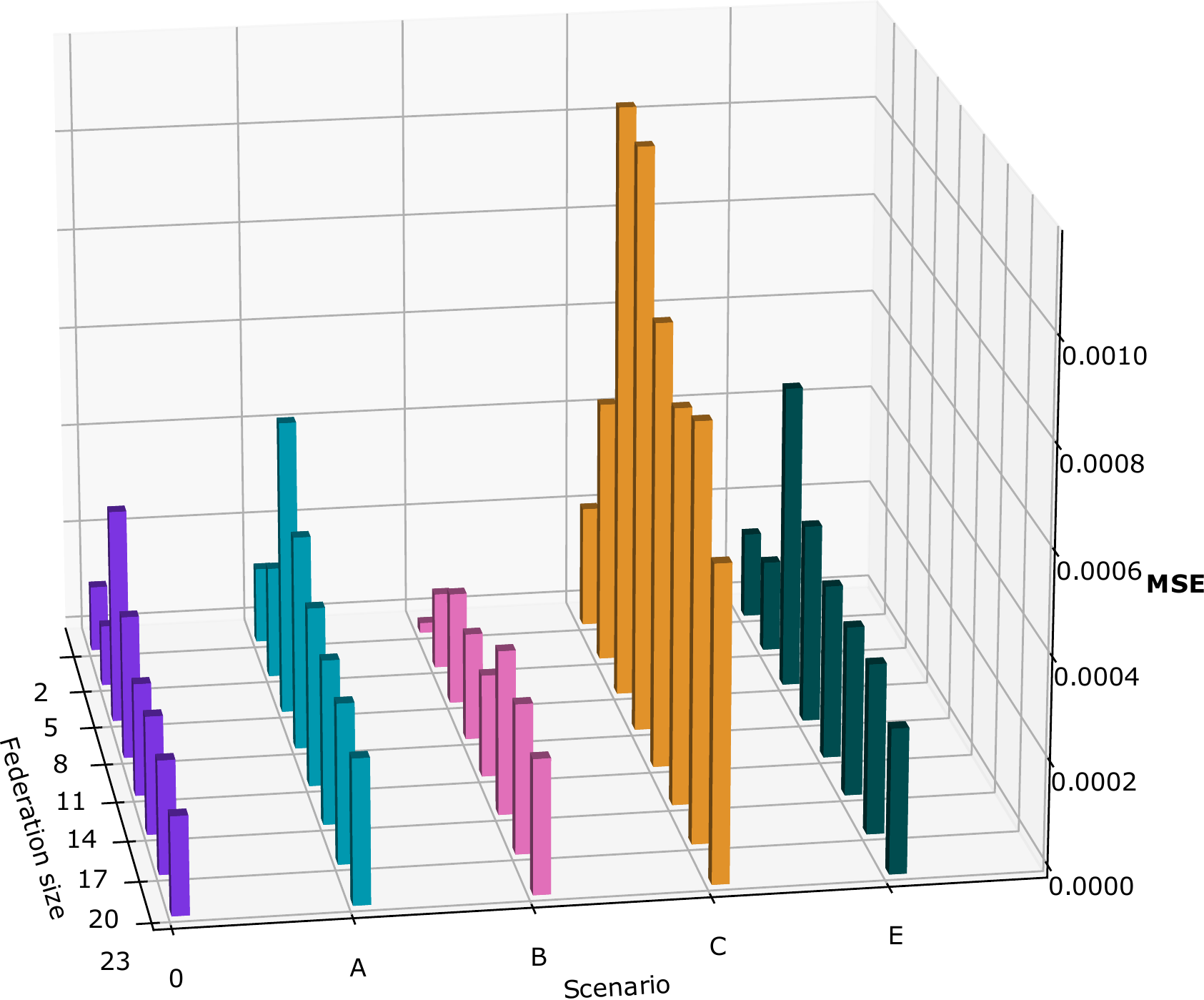}
            \caption[Network2]%
            {{\small MSE results.}}    
            \label{fig:mean and std of net14}
        \end{subfigure}
        \hfill
        \begin{subfigure}[b]{0.475\textwidth}  
            \centering 
            \includegraphics[width=\textwidth]{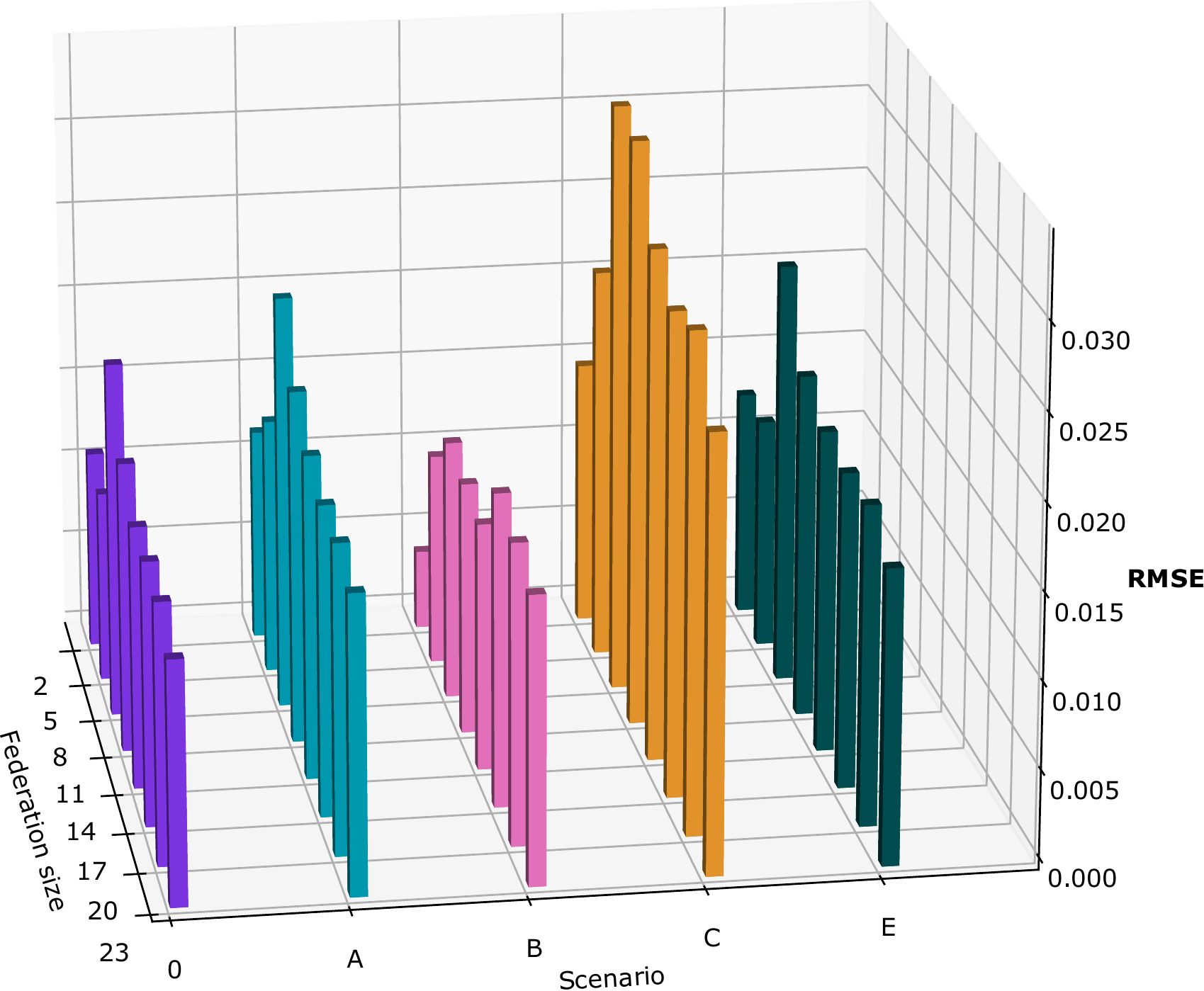}
            \caption[]%
            {{\small RMSE results.}}    
            \label{fig:mean and std of net24}
        \end{subfigure}
        \vskip\baselineskip
        \begin{subfigure}[b]{0.475\textwidth}   
            \centering 
            \includegraphics[width=\textwidth]{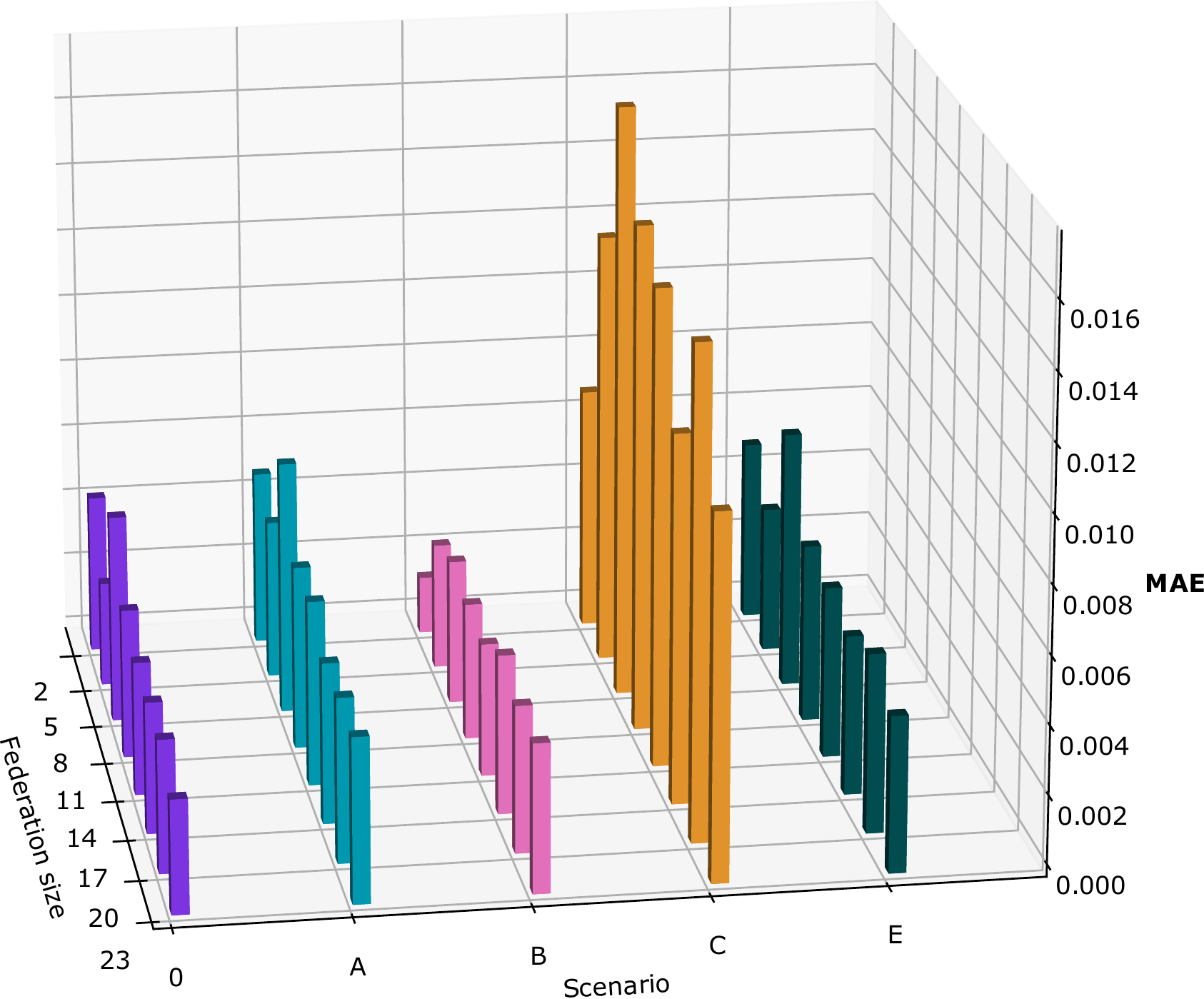}
            \caption[]%
            {{\small MAE results.}}    
            \label{fig:mean and std of net34}
        \end{subfigure}
        \hfill
        \begin{subfigure}[b]{0.475\textwidth}   
            \centering 
            \includegraphics[width=\textwidth]{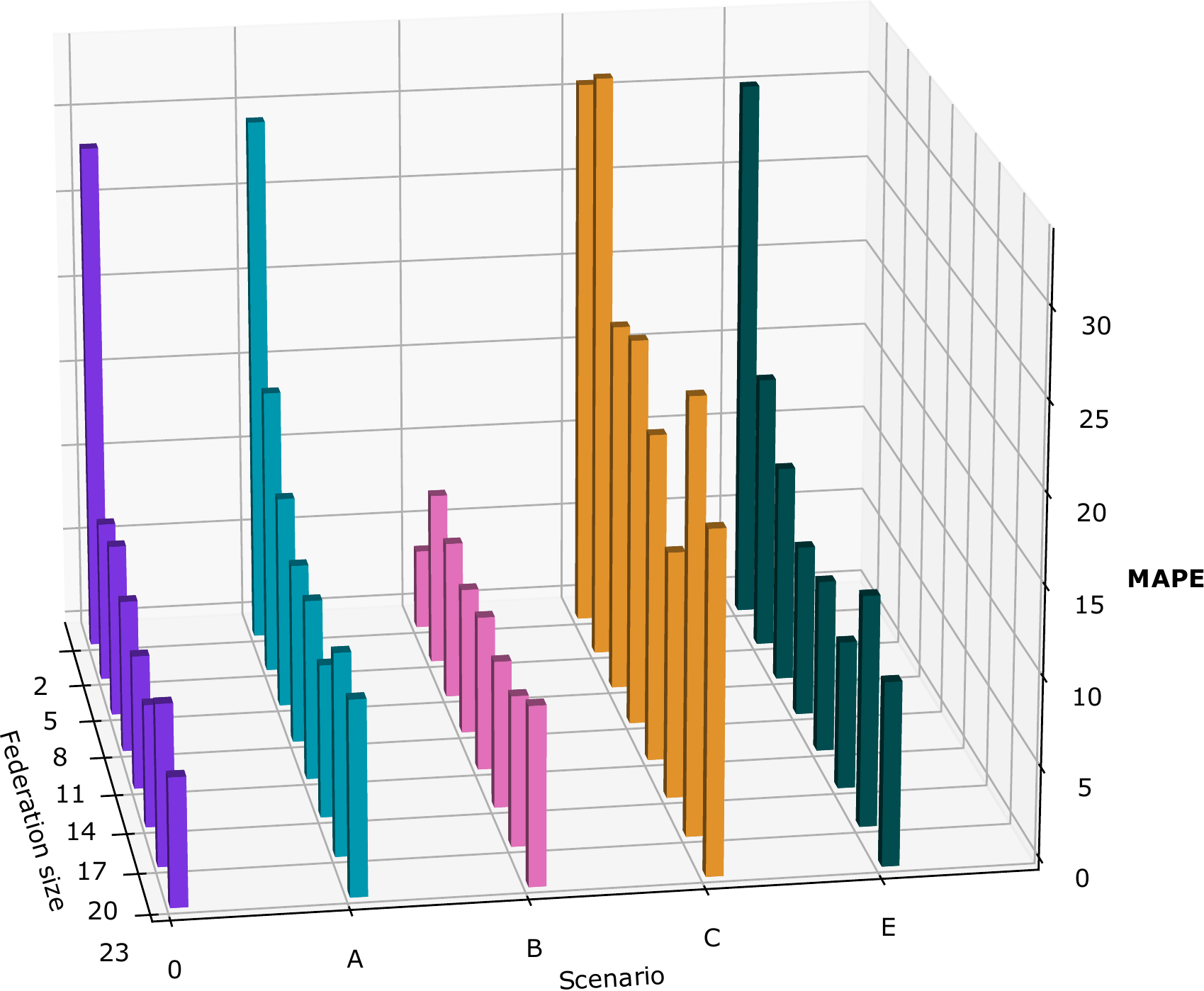}
            \caption[]%
            {{\small MAPE results.}}    
            \label{fig:mean and std of net44}
        \end{subfigure}
        \caption[]
        {\small Comparison of evaluation metrics across Scenarios 0, A, B, C, and E.} 
        \label{fig:3d_plots}
    \end{figure*}

We summarize our results for scenarios 0, A, B, C, and E in Figures \ref{fig:computation_time_comparison} and \ref{fig:3d_plots}. We omitted scenario D from these figures because in scenario D we only varied the noise scale and not the federation size.

Overall, the two figures suggest an inherent trade-off between performance and privacy in residential STLF. Yet, FL models can successfully mediate this trade-off and provide high levels of performance and privacy, especially when trained on correlated data, avoid unduly complex architectures, and employ SecAgg. 

\newpage

\section{Conclusions}
\label{sec:conclusions}

This paper analyses the use of FL and its combination with privacy preserving techniques for short-term forecasting of individual residential loads. Such a combination offers an innovative approach to accommodate both accuracy and privacy. In particular, it allows those who depend on accurate forecasts of residential loads (such as utilities, energy providers, and DSOs) to train in a collaborative fashion forecasting models with granular smart meter data without having to share this data.

Our analysis builds on historical smart meter data and consists of six scenarios. While the first two scenarios set the baseline scenarios, each of the subsequent four scenarios have a particular analytical focus. Specifically, these scenarios investigate the effects of data correlation, neural network architecture complexity, differential privacy, and secure aggregation on performance, computation time, and privacy guarantee levels. In each scenario, we also explore the effects of different federation sizes. From our analysis, we can posit the following: 

\begin{enumerate}
    \item Collaborative training of AI models with federated learning reduces forecasting accuracy as compared to a 'centralized' setting. However, it makes it easier to account for data privacy concerns through the addition of privacy-preserving techniques.
    
    \item As the number of participating clients (smart meters) in a federation increases, forecasting accuracy tends to also increase. However, while a greater number of clients leads to greater accuracy, this also implies higher computational costs that may no always be justified.
    
    \item Customer segmentation with Pearson correlation along socio-economic factors (e.g., with the ACORN methodology) substantially improves forecasting accuracy for FL models. 
    
   \item  Complex neural network architectures imply high computational costs, difficulties in handling the architecture, and a \review{potential} risk of overfitting. It is thus important to balance accuracy and usability when selecting of model architectures. 
    
    \item Complementing federated learning with differential privacy or secure aggregation does not significantly reduce forecasting accuracy but does enable very high levels of privacy.
    
    \item Adaptive and fixed clipping approaches to differential privacy provides similar performance. Adaptive clipping is easier to use as it does not require manual pre-selection of good clipping values, and it facilitates faster model convergence.
    
    \review{\item Combining autoencoder architectures with DP complicates the training of FL models. The design of these architectures magnifies the noise added by DP, which restricts the training process.}

    \item Secure aggregation is superior to DP in terms of usability, performance and computational burden. It can be added as a simple plug-and-play component, does not reduce performance by adding noise, and permits faster training.

    \end{enumerate}

Overall, our analysis suggests that a combination of federated learning with privacy-preserving techniques can be a highly promising alternative for residential short-term load forecasting. However, is not free from technical challenges. Differential privacy requires careful configuration of noise size, clipping values and client ratios to balance accuracy and privacy. Secure aggregation does not require such configuration but its cryptographic set-up can also be challenging as well. Furthermore, computational costs limit the number of clients that can be used for training.

More broadly, our study contributes to a better understanding of the use of FL and privacy-preserving techniques for residential short-term load forecasting. It makes an important contribution to the growing literature on the applications of federated learning in electric power systems by testing different NN under distributed settings, examining the implications of privacy preserving techniques, and identifying technical challenges in using FL. 

Naturally, our analysis is not free from limitations. In particular, computational costs have considerably limited the size of our federations. Even though larger federation sizes may result in somewhat different results, nevertheless we believe that our overall results are robust, as we have explored several settings in terms of: number of clients, baseline NN architectures, and dataset characteristics.

Further research may nevertheless want to (1) assess larger federation size settings with additional correlation indicators, such as the existence of distributed energy resources (i.e., photovoltaics, electric vehicles, or home energy management systems), (2) investigate data input disruptions produced by hostile agents or errors caused by malfunctions of a smart metering device, and (3) examine other, innovative NN architectures with attention mechanisms and multi-variate input data. After all, FL is highly collaborative and iterative and perfect data and operation may not always be possible in real-world applications.

\section*{Credit authorship contribution statement}
Conceptualization, J.D.F, S.P.M, C.L; Methodology, J.D.F, S.P.M, C.L; Data Curation, J.D.F, S.P.M; Writing - Original Draft, J.D.F, S.P.M, C.L;  Software J.D.F; Supervision A.R, G.F.; Writing - Review \& Editing, A.R, G.F.; Visualization, J.D.F, S.P.M.; Funding acquisition, G.F. All authors have read and agreed to the published version of the manuscript.

\section*{Declaration of Competing Interest}
The authors declare no conflict of interest.

\section*{Acknowledgements}

We would like to acknowledge Tom Josua Barberau and Orestis Papageorgiou for their valuable feedback on the first draft of this paper.

Moreover, we would like to acknowledge the support of the European Union (EU) within its Horizon 2020 programme, project MDOT (Medical Device Obligations Taskforce), Grant agreement 814654. Additionally, this project was supported by the Kopernikus-project “SynErgie” of the German Federal Ministry of Education and Research (BMBF), by PayPal and the Luxembourg National Research Fund FNR (P17/IS/13342933/PayPal-FNR/Chair in DFS/Gilbert Fridgen) as well as by the Luxembourg National Research Fund (FNR) – FiReSpARX Project, ref. 14783405.

\bibliographystyle{IEEEtranN}
\bibliography{0_main}  

\begin{thebibliography}{10}
\bibitem{hao2019training}
K.~Hao, Training a single ai model can emit as much carbon as five cars in
  their lifetimes, MIT Technology Review.
\end{thebibliography}






\end{document}